\documentclass[letterpaper, 10 pt, conference]{ieeeconf}  % Comment this line out if you need a4paper

\IEEEoverridecommandlockouts 
\usepackage{graphics} % for pdf, bitmapped graphics files
\usepackage{epsfig} % for postscript graphics files
\usepackage{mathptmx} % assumes new font selection scheme installed
\usepackage{times} % assumes new font selection scheme installed
\usepackage{amsmath} % assumes amsmath package installed
\usepackage{amssymb}  % assumes amsmath package installed
\usepackage{listings}
\usepackage{hyperref}
\usepackage{xcolor}
\usepackage{booktabs}

\def\Snospace~{\S{}}

\let\citet=\cite

\title{\LARGE \bf
Toward Grounded Commonsense Reasoning}

\lstdefinestyle{mystyle}{
    backgroundcolor=\color{backcolour},   
    commentstyle=\color{codepurple},
    keywordstyle=\color{magenta},
    numberstyle=\tiny\color{codegray},
    stringstyle=\color{codegreen},
    basicstyle=\ttfamily\footnotesize,
    breakatwhitespace=false,         
    breaklines=true,                 
    captionpos=b,                    
    keepspaces=true,                 
    numbers=left,                    
    numbersep=5pt,                  
    showspaces=false,                
    showstringspaces=false,
    showtabs=false,                  
    tabsize=2
}
\definecolor[named]{xILIADOrange}{HTML}{FF9B00}
\definecolor[named]{xRefBlue}{HTML}{0055C9}

\definecolor[named]{pOurPink}{HTML}{DE006F}
\definecolor[named]{pAblationLightPink}{HTML}{FF41A0}
\definecolor[named]{pBaselineGray}{HTML}{5A5A5A}
\definecolor[named]{pWeakLightGray}{HTML}{C0C0C0}

\definecolor[named]{cMinae}{HTML}{B70101}
\definecolor[named]{cVivek}{HTML}{D027D9}
\definecolor[named]{cSidd}{HTML}{18647E}
\definecolor[named]{cHengyuan}{HTML}{DCF763}

\definecolor[named]{cAnca}{HTML}{FF0000}
\definecolor[named]{cDorsa}{HTML}{FF9D0A}

\definecolor{codegreen}{rgb}{0,0.6,0}
\definecolor{codegray}{rgb}{0.5,0.5,0.5}
\definecolor{codepurple}{rgb}{0.58,0,0.82}
\definecolor{backcolour}{rgb}{0.95,0.95,0.92}

\newcommand{\dataset}{\textsc{MessySurfaces}}

\hypersetup{
    colorlinks=true,
    linkcolor=xILIADOrange,
    filecolor=magenta,
    urlcolor=xILIADOrange,
    citecolor=xILIADOrange,
}

\let\citep=\cite

\author{Minae Kwon,\ \ Hengyuan Hu,\ \ Vivek Myers$^\dagger$,\ \ Siddharth Karamcheti,\ \ Anca Dragan$^\dagger$,\ \ Dorsa Sadigh  \\
Stanford University, \ \ UC Berkeley$^\dagger$\\
\{\texttt{mnkwon}, \texttt{hengyuan}, \texttt{skaramcheti}, \texttt{dorsa}\}\texttt{@cs.stanford.edu}, \\  \{\texttt{vmyers}, \texttt{anca}\}\texttt{@berkeley.edu}$^\dagger$%
}

\makeatletter
\let\@old@paragraph\paragraph
\def\@strip@dot#1.\@end{#1}
\def\paragraph#1{\@old@paragraph{\@strip@dot#1\@end}}
\makeatother

\begin{document}

\maketitle
\thispagestyle{empty}
\pagestyle{empty}

\begin{abstract}
Consider a robot tasked with tidying a desk with a meticulously constructed Lego sports car. A human may recognize that it is not appropriate to disassemble the sports car and put it away as part of the ``tidying.'' How can a robot reach that conclusion? Although large language models (LLMs) have recently been used to enable commonsense reasoning, grounding this reasoning in the real world has been challenging. To reason in the real world, robots must go beyond passively querying LLMs and \emph{actively gather information from the environment} that is required to make the right decision. For instance, after detecting that there is an occluded car, the robot may need to actively perceive the car to know whether it is an advanced model car made out of Legos or a toy car built by a toddler. We propose an approach that leverages an LLM and vision language model (VLM) to help a robot actively perceive its environment to perform grounded commonsense reasoning. To evaluate our framework at scale, we release the \dataset{} dataset which contains images of $70$ real-world surfaces that need to be cleaned. We additionally illustrate our approach with a robot on $2$ carefully designed surfaces. We find an average $12.9\%$ improvement on the \dataset{} benchmark and an average $15\%$ improvement on the robot experiments over baselines that do not use active perception. The dataset, code, and videos of our approach can be found at \url{https://minaek.github.io/grounded_commonsense_reasoning/}.
\end{abstract}

\section{Introduction}
\label{sec:introduction}

Imagine you are asked to clean up a desk and you see a meticulously constructed Lego sports car on it. You might immediately recognize that the normative behavior is to leave the car be, rather than taking it apart and putting it away as part of the ``cleaning''. But how would a robot in that same position know that's the right thing to do? Traditionally, we would expect this information to be specified in the robot's objective -- either learned from demonstrations \citep{ross2011reduction,brownBetterthanDemonstratorImitationLearning2019,palanLearningRewardFunctions2019} or from human feedback \citep{sadigh2017active,liLearningHumanObjectives2021,biyikLearningRewardFunctions2021,fitzgeraldINQUIREINteractiveQuerying}. While a robot could expensively query a human for their preferences on how to clean the car, we explore a different question in this work: how can we equip robots with the commonsense reasoning necessary to follow normative behavior \textit{in the absence of personalized input from the human}? The ability to behave in a commonsense, normative manner can be an effective prior over robot behavior when personalized feedback is not present. When feedback is present, having a good prior can reduce the amount of human specification needed.

\begin{figure*}
    \centering
    \includegraphics[width=\textwidth]{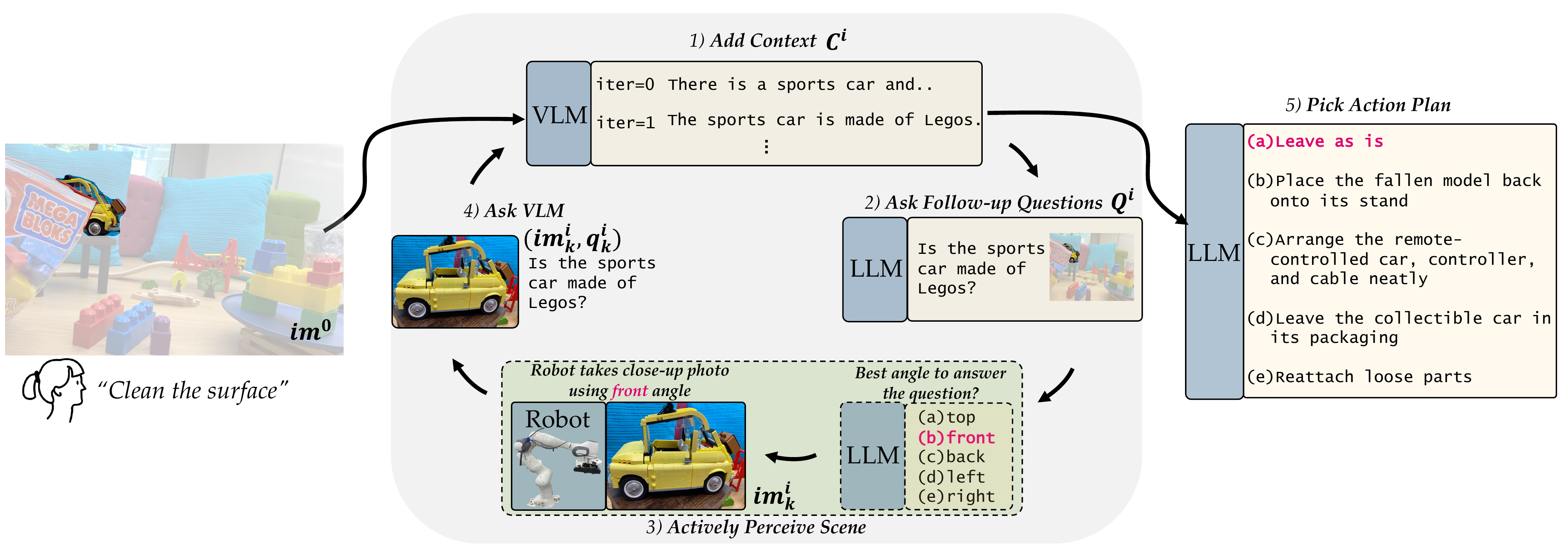}
    \caption{\small \textbf{Grounded Commonsense Reasoning Framework.} We demonstrate our framework using the sports car. Blue boxes indicate the model and yellow boxes indicate its output. 
    % The scene image is blurred for clarity. 
    Our framework takes an 
    % (unblurred) 
    image of the scene and an instruction as input. 1) The VLM outputs an initial description of the scene $\mathcal{C}^0$ from the initial image $im^0$. 2) The LLM asks follow-up questions about each object in the scene, $\mathcal{Q}^i$. 3) The robot takes a close-up image $im^i_k$ of each object $k$. It is guided by an LLM that chooses the best angle that would help answer the question. 4) We pair the close-up images with the follow-up questions and ask the VLM to answer them. Answers are appended to the context. We repeat steps 1-4 to gather more information. 5) We query an LLM to choose the most appropriate way to tidy the object.}
    \label{fig:frontfig}
    \vspace{-1.5 em}
\end{figure*}

%However, Lego sports cars are not common, and it is challenging for humans to specify a priori what objects a robot might encounter \citep{amodei2016concrete,hadfield-menellInverseRewardDesign2017}. While a robot could expensively query a human for what to do during these circumstances, we explore a different question in this work: \textit{how can we enrich robots with the commonsense reasoning necessary to know what to do, without human intervention?}

Recent work has demonstrated that large language models (LLMs) trained on internet data have enough context for commonsense reasoning \citep{talmor2022commonsenseqa}, making moral judgements \citep{jiang2021can,hendrycks2020aligning}, or acting as a proxy reward function capturing human preferences~\cite{kwon2023reward}. Rather than explicitly asking a human for the answer, the robot could instead ask an LLM whether it would be appropriate to clean up the car. But in real-world environments, this is easier said than done. Tapping into an LLM's commonsense reasoning skills in the real-world requires the ability to \textit{ground language in the robot's perception of the world} -- an ability that might be afforded by powerful vision-and-language models (VLMs). Unfortunately, we find that today's VLMs cannot reliably provide all the relevant information for commonsense reasoning. For instance, a VLM may not describe that the sports car is constructed from Legos, or that it contains over $1000$ pieces -- details that are key to making decisions. While advanced multi-modal models might alleviate this problem, a fundamental limitation is the image itself might not contain all the relevant information. If the sports car is partially occluded by a bag (as in \autoref{fig:frontfig}), no VLM could provide the necessary context for reasoning over what actions to take. Such a system would instead need the ability to move the bag -- or move \textit{itself} -- to actively gather the necessary information. Thus, in order to perform ``grounded commonsense reasoning'' robots must go beyond passively querying LLMs and VLMs to obtain action plans and instead \emph{directly interact with the environment.} Our insight is that robots must reason about what additional information they need to make appropriate decisions, \textit{and then actively perceive the environment to gather that information}.

Acting on this insight, we propose a framework to enable a robot to perform grounded commonsense reasoning by 
iteratively identifying details it still needs to clarify about the scene before it can make a decision (e.g. is the model car made out of intricate Lego pieces or MEGA Bloks?) and actively gathering new observations to help answer those questions (e.g. getting a close up of the car from a better angle). In this paper, we focus on the task of cleaning up real-world surfaces through commonsense reasoning. Our framework is shown in \autoref{fig:frontfig}. Given a textual description of the desk, an LLM asks follow-up questions about the state of each object that it needs in order to make a decision of what the robot should do with that object. The robot actively perceives the scene by taking close-up photos of each object from angles suggested by the LLM. The follow-up questions and close-up photos are then given to a VLM so that it can provide more information about the scene. 

This process can be repeated multiple times. The LLM then decides on an action the robot should take to clean the object in an appropriate manner. For example, our robot leaves the Lego sports car intact, throws a browning half-eaten banana in the trash, but keeps an unopened can of Yerba Mate on the desk. Furthermore, we release the \dataset{} dataset containing images of $70$ surfaces as well an evaluation benchmark that assesses how well a robot can clean up each surface in an appropriate manner. The dataset is available \href{https://sites.google.com/view/social-grounding}{here}. 
%\href{https://minaek.github.io/groundedsocialreasoning/}{here}.

We evaluate our framework on our benchmark dataset as well as on a real-world robotic system. We examine each component of our framework, asking whether the robot asks useful follow-up questions, whether the robot chooses informative close-up images, and whether the images actually help a VLM more accurately answer questions. We find an average $12.9\%$ improvement on the \dataset{} benchmark and an average $15\%$ improvement on the robot experiments over baselines that do not use active perception.

% \begin{figure*}
%     \centering
%     \includegraphics[width=\textwidth]{figures/front_fig.pdf}
%     \caption{\small \textbf{Grounded Commonsense Reasoning Framework.} We demonstrate our framework using the sports car. Blue boxes indicate the model and yellow boxes indicate its output. 
%     % The scene image is blurred for clarity. 
%     Our framework takes an 
%     % (unblurred) 
%     image of the scene and an instruction as input. 1) The VLM outputs an initial description of the scene $\mathcal{C}^0$ from the initial image $im^0$. 2) The LLM asks follow-up questions about each object in the scene, $\mathcal{Q}^i$. 3) The robot takes a close-up image $im^i_k$ of each object $k$. It is guided by an LLM that chooses the best angle that would help answer the question. 4) We pair the close-up images with the follow-up questions and ask the VLM to answer them. Answers are appended to the context. We repeat steps 1-4 to gather more information. 5) Finally, we query an LLM to choose the most appropriate way to tidy the object.}
%     \label{fig:frontfig}
%     \vspace{-1.5 em}
% \end{figure*}

\section{Related Work}
\label{sec:related-work}
% === Related Work ===
\paragraph{Commonsense Reasoning.}
Large language models are trained on internet-scale data, making them effective commonsense reasoners \citep{brown2020gpt3,ryttingLeveragingInductiveBias,zhangLargeLanguageModels2023,zhouEvaluatingCommonsensePreTrained2020}. 
Prior works have studied whether LLMs' commonsense reasoning aligns with human values \citep{jiang2021can,hendrycks2020aligning,jinWhenMakeExceptions2022,kwon2023reward}.  
There is evidence that when LLMs make moral or social judgements, they align with the normative beliefs of the population that generated their training data \citep{fraserDoesMoralCode2022}. In addition, prior work show commonsense reasoning models can align with conventional beliefs \citep{ammanabroluAligningSocialNorms2022,hendrycksWhatWouldJiminy2022,hendrycksAligningAIShared2023,zellersRecognitionCognitionVisual2019}. %Researchers have raised concerns that amplifying normative beliefs within biased training data populations could disproportionately harm minority groups \cite{touilebMeasuringNormativeDescriptive2023a,blodgettLanguageTechnologyPower2020,hooverBoundHatredRole2019}.
\textit{Our approach is in line with commonsense reasoning; instead of adapting to individual preferences, we show we can take commonsense actions to clean up a scene.}

\paragraph{Learning Human Preferences.}
Past work on aligning with human preferences has focused on using human feedback to infer rewards and policies by designing queries for active preference learning \citep{akrourAprilActivePreference2012,sadigh2017active,biyikLearningRewardFunctions2021,cakmakHumanPreferencesRobothuman2011}, performing inverse reinforcement learning \citep{ziebartMaximumEntropyInverse2008,ratliffMaximumMarginPlanning2006}, or recovering reward signals from language feedback \citep{kwon2023reward,fanMineDojoBuildingOpenEnded2022,singhConcept2RobotImprovingLearning,shaoConcept2RobotLearningManipulation2021,mirchandani2021ella}. Policies defined via LLMs have also been directly tuned with language feedback by approaches like RLHF \citep{zieglerFineTuningLanguageModels2020}. Instead of querying humans, we leverage normative values from pre-trained models. While some works use normative values from LLMs in negotiations and games \citep{hu2023language}, these are not grounded in the real world. \textit{In this work, we do not focus on particular human preferences, though the normative responses of LLMs could be fine-tuned for particular applications.}

\paragraph{Active Perception.}
%Within the brain, action and perception are intimately linked \cite{bajcsyActivePerception1988,pulvermullerActivePerceptionSensorimotor2010}. 
%Biological agents manipulate their environment to select and gate their perceptual information, which is in turn processed to select further actions. 
%Although essential to natural intelligence, current AI approaches often do not close this loop between action and perception \cite{bajcsyRevisitingActivePerception2018,bohgInteractivePerceptionLeveraging2017}. 
%\vm{TidyBot is very related work and we should be clear to distinguish our active perception from theirs.}
When robots must use commonsense reasoning like humans, active information gathering may be important \citep{bohgInteractivePerceptionLeveraging2017}. % to behaving in a way consistent with human values. 
Approaches like TidyBot actively zoom-in on objects to better categorize them \citep{wuTidyBotPersonalizedRobot2023}. 
%However, perception is not truly active in this approach---different shot angles may be needed to gather information about objects. 
Other approaches such as Inner Monologue seek out additional environment information, but need aid from a human annotator or assume access to simulators \citep{huangInnerMonologueEmbodied2022,zhaoChatEnvironmentInteractive2023}. %In simulation, language models can perform active perception to learn physical object attributes \cite{zhaoChatEnvironmentInteractive2023}. 
%\vm{This last work \cite{zhaoChatEnvironmentInteractive2023} is superficially very similar to what we are doing; we may want to compare more.} 
VLMs have also been used for active perception in navigation \citep{yu2022parti,huangVisualLanguageMaps2023,shahLMNavRoboticNavigation2022}. \textit{In this work, we show that active perception is necessary for grounded commonsense reasoning, enabled by the semantic knowledge in an LLM.}

\paragraph{LLMs for Robotics.}
Past work uses semantic knowledge in LLMs for task planning. 
Methods like SayCan decompose natural language tasks into primitive action plans \citep{ahnCanNotSay2022,attarianSeePlanPredict2022,Huang2022LanguageMA}. 
%Benchmarks like ALFRED can be used to evaluate the performance of planning and task composition with natural language \cite{shridharALFREDBenchmarkInterpreting2020,meesCALVINBenchmarkLanguageConditioned2022}. 
%Meanwhile, approaches like TidyBot \cite{wuTidyBotPersonalizedRobot2023} use example tasks specified with language to infer human preferences about how to manipulate a scene.
% \vm{already mentioned above}
In addition, approaches such as Code as Policies \citep{liangCodePoliciesLanguage2023,surisViperGPTVisualInference2023} use LLMs to write Python programs that plan with executable robot policy code. %VLMs play an important role to ground the reasoning of this body of work \cite{radford2021clip,liuVisualInstructionTuning2023,cuiCanFoundationModels2022a,shahLMNavRoboticNavigation2022,Huang2022LanguageMA}.
%Other approaches use learned vision-language representations of tasks to ground instructions \cite{lynchLanguageConditionedImitation2021,jangBCZZeroShotTask2021,cuiCanFoundationModels2022a,shahLMNavRoboticNavigation2022a,shahGNMGeneralNavigation2022}.  
%Alternatively, models can recover reward signals from language \cite{kwon2023reward,fanMineDojoBuildingOpenEnded2022,singhConcept2RobotImprovingLearning,shaoConcept2RobotLearningManipulation2021,mirchandani2021ella}. 
%Direct vision-language grounding as tasks or rewards has the advantage of enabling end-to-end learning of low-level control without predefined action or perception primitives.
Other approaches use multimodal sequence models to reason about language-conditioned manipulation \citep{brohanRT1RoboticsTransformer2022,driessPaLMEEmbodiedMultimodal2023,reedGeneralistAgent2022,jiangVIMAGeneralRobot2022}. 
 %These methods excel at multimodal reasoning in embodied settings, but are limited by the large amounts of data needed for training.
 \textit{We use the semantic awareness of an LLM to reason about action plans. Unlike the above works, an LLM interactively queries an off-the-shelf VLM to ground the scene.}

\section{Grounding Commonsense Reasoning}
\label{sec:framework}
% === Framework ===

We propose a framework that combines existing foundation models in a novel way to enable active information gathering, shown in \autoref{fig:frontfig}. Our framework makes multiple calls to an LLM and VLM to gather information. The LLM plays a number of distinct roles in our framework that we distinguish below: generating informative follow-up questions, guiding active perception, and choosing an action plan. In every call, the LLM takes in and outputs a string $\texttt{LLM}: A^* \rightarrow A^*$, and the VLM takes in an image, string pair and outputs a string $\texttt{VLM}: \mathcal{I} \times A^* \rightarrow A^*$, where $A^*$ is the set of all strings and $\mathcal{I}$ is the set of all images. The context $\mathcal{C}^i \in A^*$ contains information about the scene that the robot has gathered up to iteration $i$ of the framework. Initially, the inputs to our framework are an image of the scene $im^0 \in \mathcal{I}$ (i.e., an unblurred image from \autoref{fig:frontfig}) and an instruction (e.g., ``clean the surface''). 

\noindent \textbf{VLM Describes the Scene.} Our framework starts with the VLM producing an initial description $\mathcal{C}^0$ of the scene from the scene image $im^0$. The description can contain varying amounts of information --- in the most uninformative case, it may simply list the objects that are present. In our experiments, this is the description that we use. %We assume that the initial description is accurate to fairly evaluate the rest of our approach.

\noindent \textbf{LLM Generates Follow-Up Questions.} To identify what information is missing from $\mathcal{C}^0$, we use an LLM to generate informative follow-up questions as shown in stage (2) of \autoref{fig:frontfig}. We prompt an LLM with $\mathcal{C}^0$ and ask the LLM to produce a set of follow-up questions $\mathcal{Q}^i = \{q^i_1, \ldots, q^i_K\}$ for the $K$ objects. LLMs are apt for this task because of their commonsense reasoning abilities. We use Chain-of-Thought prompting \citep{wei2022chain} where we first ask the LLM to reason about the appropriate way to tidy each object before producing a follow-up question (see examples in the supplementary). For example, the LLM could reason that the sports car should be put away if it is a toy but left on display if someone built it. The resulting follow-up question asks whether the sports car is built with Lego blocks. We assume that the information in $\mathcal{C}^0$ is accurate (i.e., correctly lists the names of all the objects) to prevent the LLM from generating questions based on inaccurate information.

\noindent \textbf{Robot Actively Perceives the Scene.} At this stage, one might normally query the VLM with the original scene image $im^0$. However if the object-in-question is obstructed or too small to see, the scene image might not provide enough information for the VLM to answer the follow-up question accurately (e.g., the sports car is obstructed in \autoref{fig:frontfig}). Instead, we would like to provide an unobstructed close-up image $im_k^i \in \mathcal{I}$ of the object $k$ to ``help'' the VLM accurately answer the generated questions. Taking informative close-up images requires interaction with the environment --- something we can use a robot for. 

To actively gather information, the robot should proceed based on some notion of ``informativeness'' of camera angles. To determine ``informativeness'', we can again rely on the commonsense knowledge of LLMs. Although LLMs don't have detailed visual information about the object, they can suggest reasonable angles that will be, on average, informative. For instance, an LLM will choose to take a photo from the top of an opaque mug, instead of its sides, to see its content. In practice, we find that this approach works well and can improve the informativeness of an image by $8\%$. We query an LLM to choose a close-up angle of the object from a set of angles $\{\texttt{<FRONT>, <BACK>, <LEFT>, <RIGHT>, <TOP>}\}$ that would give an unobstructed view. We then pair the close-up images with their questions $\{(im^i_1, q^i_1), \ldots, (im^i_k, q^i_K)\}$ and query the VLM for answers to these questions in step (4) of our framework. We concatenate the VLM's answers for each object and append them to our context $\mathcal{C}^i$ to complete the iteration. To gather more information about each object, steps $1-4$ can be repeated where the number of iterations is a tunable parameter. 

\noindent \textbf{LLM Chooses an Action Plan.} In the final step, for each object, we prompt the LLM with the context $\mathcal{C}^i$ and a multiple choice question that lists different ways to tidy an object. The LLM is then instructed to choose the most appropriate option. The multiple choice options come from the \dataset{} benchmark questions, a bank of $308$ multiple-choice questions about how to clean up real-life objects found on messy surfaces. For example, in \autoref{fig:frontfig}, the LLM chooses to leave the sports car as is because it infers that the sports car must be on display. To map the natural language action to robot behavior, we implement a series of hand-coded programmatic skill primitives that define an API the LLM can call into. See \autoref{sec:experiments} for more details. 

%We evaluate this step using the \dataset{} benchmark questions,  where we query the LLM to choose one of the multiple choice options. These components --- asking questions, actively perceiving the environment, using a VLM to answer our questions, and choosing an appropriate action plan --- form the bases of our approach.

\begin{figure*}
    \centering
    \includegraphics[scale=0.4]{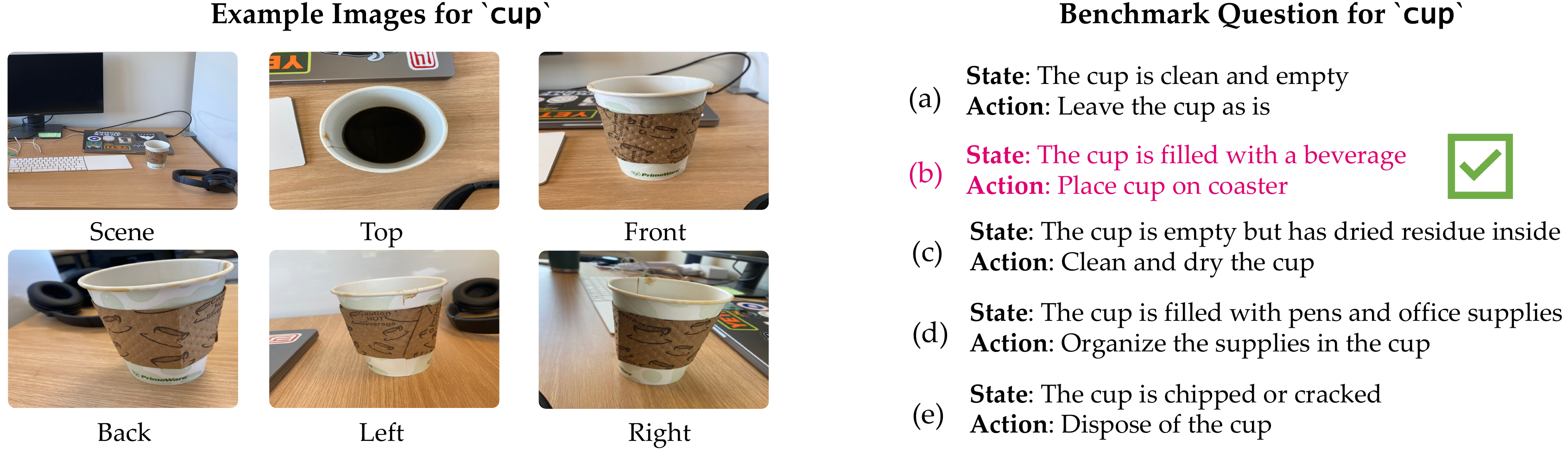}
    \caption{\small \textbf{\dataset{} Example.} Each object in \dataset{} is represented by a scene image and $5$ close-up images. Each object also has a benchmark question that presents $5$ options to tidy the object; each option is constructed by producing a cleaning action conditioned on a hypothetical object state.}
    \label{fig:benchmark_example}
\end{figure*}

\section{The \dataset{} Dataset}
\label{sec:dataset}
% === Dataset ===

% === Body Text ===

To assess a robot's ability to perform commonsense reasoning in grounded environments, we introduce the \dataset{} dataset. The dataset consists of images of $308$ objects across $70$ real-world surfaces that need to be cleaned. %The images consist of both scene-level and close-up images of each object. 
An average of $68\%$ of objects are occluded in scene-level images\footnote{Computed as the average number of times annotators indicated a question cannot be answered by the scene image.}, so we also provide $5$ close-up images as a way for the robot to ``actively perceive'' the object, see \autoref{fig:benchmark_example} for an example. \dataset{} also includes a benchmark evaluation of multiple choice questions for each object where each option corresponds to different ways to tidy the object. Through a consensus of $5$ human annotators, we determine which one of the choices is the most  appropriate. To do well, a robot must reason about the appropriate way to clean each object from the images alone. Since no human preferences are given, the robot must identify relevant attributes of each object from the images (e.g., is the sports car built out of Legos or MEGA Bloks?) and then reason about how to tidy the object using this information. \dataset{} contains $45$ office desks, $4$ bathroom counters, $5$ bedroom tables, $8$ kitchen counters, $4$ living room tables and $4$ dining tables.  %LOL it's not balanced at all :D $

% Should discuss whether we want to release our follow-up questions as part of this dataset

\paragraph{Data Collection Process.}
We recruited $51$ participants to provide images of cluttered surfaces. Each participant was asked to pick $4$ -- $6$ objects on a surface. They were then asked to take a photo of the scene-level view as well as close-up photos of each object from the top, right, left, front, and back angles -- the offline equivalent of having a robot actively navigate a scene. The task took approximately $15-30$ minutes. After receiving the photos, we post-processed each image and cropped out any identifiable information. 

\paragraph{Benchmark Evaluation.}
The benchmark questions consist of $5$ LLM-generated multiple choice options about how to manipulate each object to clean the surface in an appropriate manner. To make the options diverse, we asked the LLM to first identify $5$ states the object could be in and then queried it to come up with a cleaning action for each of those states (see \autoref{fig:benchmark_example} for an example). For each question, we recruited $5$ annotators to choose the correct state-action pair based on the scene and close-up images of the object. Annotators were also given an option to indicate if none of the choices were a good fit. We used the majority label as our answer and omitted $16$ questions (out of $324$) where a majority thought none of the choices were a good fit. For questions that had two equally popular answers, we counted both as correct. Our annotators agreed on average $67\%$ of the time. To evaluate the quality of our multiple choice options, we asked annotators to rate how appropriate each cleaning action is for each object state. Annotators gave each option an average rating of $4.1$ out of $5$. The average rating for the correct option was $4.4$ out of $5$. \textit{Annotators.} In total, we recruited $350$ English-speaking annotators from Prolific that were based in the U.S.\ or U.K.\ with an approval rating of at least $98\%$. Our study is IRB-approved.

\section{Experiments}
\label{sec:experiments}
% === Experiments ===

% \begin{figure*}
%     \centering
%     \includegraphics[width=\textwidth]{figures/benchmark.pdf}
%     \caption{\small \textbf{\dataset{} Benchmark Accuracy.} For both the Oracle VLM and InstructBLIP, on average, our approach outperforms all baselines on the \dataset{} benchmark. Accuracy is given by the percentage by which our framework selects the most appropriate (as indicated by our annotators) way to tidy each object.} %Since there are $5$ options, randomly selecting an answer would give $20\%$ accuracy.}
%     \label{fig:benchmark}
% \end{figure*}

\begin{figure}
    \centering
    \includegraphics[width=\linewidth]{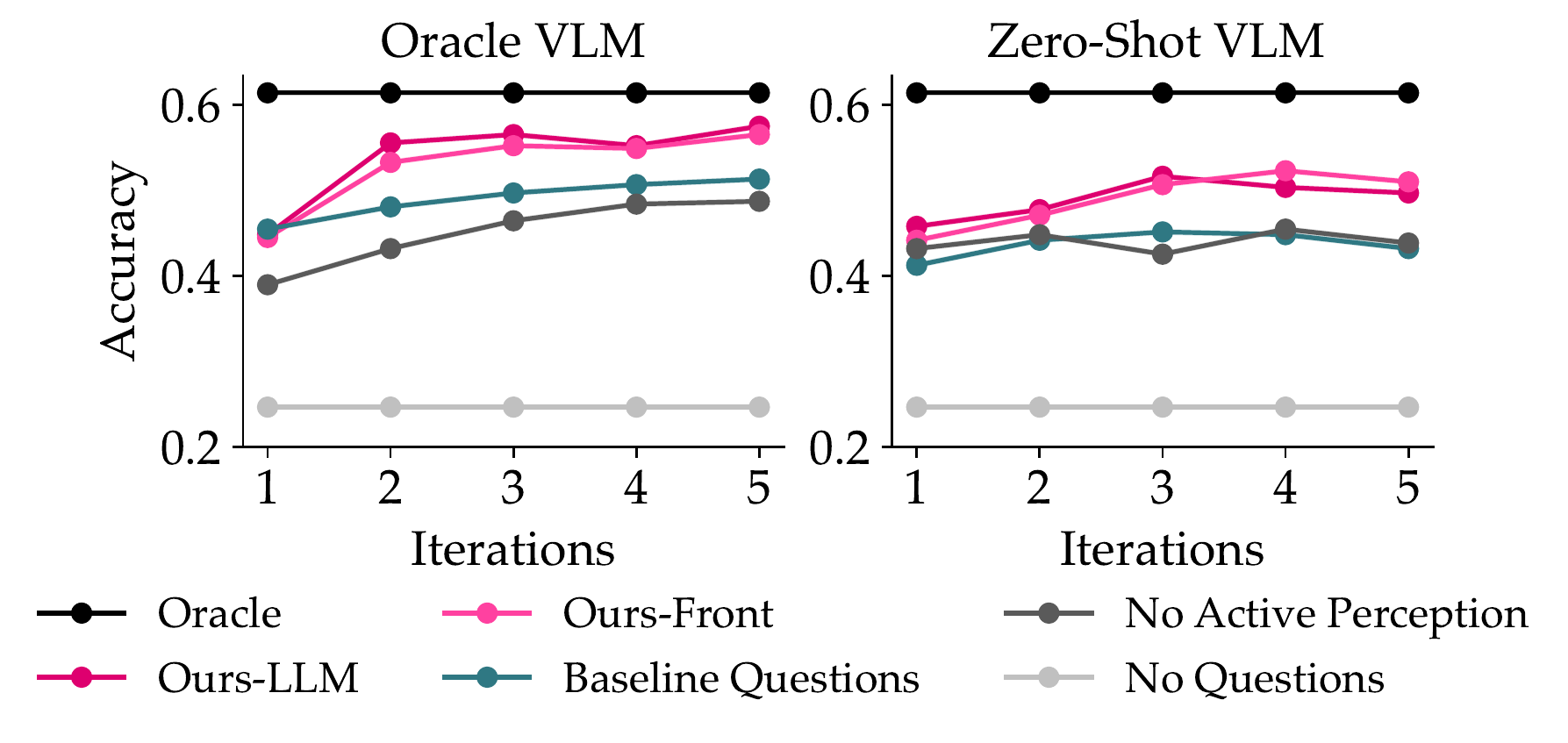}
    \caption{\small \textbf{\dataset{} Benchmark Accuracy.} For both the Oracle VLM and InstructBLIP, on average, our approach outperforms all baselines on the \dataset{} benchmark. Accuracy is given by the percentage by which our framework selects the most appropriate (as indicated by our annotators) way to tidy each object.} %Since there are $5$ options, randomly selecting an answer would give $20\%$ accuracy.}
    \label{fig:benchmark}
\end{figure}

% === Body Text ===

We examine how well our approach can perform grounded commonsense reasoning on the \dataset{} dataset as well as a real-world robotic system. 

\noindent{\textbf{Primary Metric.}} We use accuracy on the benchmark questions as our primary metric. Each benchmark question presents $5$ options on how to tidy the object, with accuracy defined as the percentage by which our framework selects the most appropriate option (as indicated by our annotators).

\newcommand{\legend}[1]{\raisebox{2pt}{\fcolorbox{black}[HTML]{#1}{\rule{0pt}{2.5pt}\rule{2.5pt}{0pt}}}}

\noindent \textbf{Baselines.} Key to our approach (\legend{DE006F} \textbf{Ours-LLM}) is the ability to supplement missing information by asking questions and actively perceiving the environment. To evaluate this, we compare the following:
\begin{itemize}
    \item \legend{000000} \textbf{Oracle.} We ask a human annotator to answer the benchmark questions where they can actively perceive the scene using all angles.
    \item \legend{DE006F} \textbf{Ours-LLM.} Our approach as described in \autoref{sec:framework}.
    \item \legend{FF41A0} \textbf{Ours - Front.} Inspired by TidyBot~\cite{wuTidyBotPersonalizedRobot2023}, this is a variant of our approach wherein we simulate ``zooming'' into the image, using the ``front'' angle image as input to the VLM. The ``front'' angles can be the most informative angle in many cases, making it an effective heuristic.
    \item \legend{2f7883} \textbf{Baseline Questions.} This baseline evaluates the need for normative commonsense reasoning by asking more  factual questions (e.g., ``What color is the cup?'').
    \item \legend{5A5A5A} \textbf{No Active Perception.} This baseline evaluates the need for active perception in our framework by allowing the robot to ask questions \textit{that are answered solely from the scene image}.
    \item \legend{C0C0C0} \textbf{No Questions.} This baseline requires the robot to perform grounded commonsense reasoning from an initial description of the scene. The robot does not ask questions or actively perceive the environment, instead operating in an open-loop fashion akin to methods like SayCan \citep{ahnCanNotSay2022}. 
\end{itemize}

\noindent \textbf{Implementation Details.} We use GPT-4 with temperature $0$ as our LLM and InstructBLIP \citep{dai2023instructblip} (Flan-T5-XXL) as our VLM. We also report ``oracle'' results where a human answers questions instead of the VLM to simulate results our approach could achieve if the VLM were near-perfect (denoted as the ``Oracle VLM''). Further implementation details (e.g., prompts, model usage) are in the supplementary.

\subsection{Evaluation on \dataset{}}

% Evaluating on \dataset{} allows us to test our framework at scale on realistic surfaces. In addition to reporting benchmark question accuracy, we evaluate each component of our framework independently. We employ the same set of $350$ Prolific annotators described in \autoref{sec:dataset} to perform our analyses. 

% \noindent \textbf{Benchmark Evaluation Results.} 

We evaluate our method on the $308$ benchmark questions across $5$ iterations of our framework. After each iteration, the robot is evaluated on the information it has accumulated up until that point. We measure accuracy on each question and report results using both the Oracle VLM and zero-shot performance on InstructBLIP. Although \textbf{No Question} and \textbf{Oracle} are ``open-loop'' methods that do not require iteration, we plot their results as a constant for comparison. 

%We measure accuracy on each question as described in~\cref{sec:dataset}. Since the robot accumulates more information at each iteration, we expected the benchmark accuracy to monotonically improve where accuracy at iteration $5$ would be the highest. We find that accuracy stops improving significantly after iteration $2$, perhaps due to the waning quality of follow-up questions, see~\cref{fig:question}. Although \textbf{No Question} and \textbf{Oracle} are ``open-loop'' methods that do not require iteration, we plot their accuracies as a constant across iterations for comparison. 

\textbf{After $5$ iterations, for both the Oracle VLM and InstructBLIP, our approaches outperform all baselines}: \textbf{No Question}, \textbf{No Active Perception}, and \textbf{Baseline Questions}. Notably, \textbf{Ours-LLM} significantly outperforms \textbf{No Question} by an average of $27.7\%$ across the two VLM types, $p<0.01$. \textbf{Ours-LLM} also outperforms \textbf{Baseline Questions} by an average of $5\%$ across the VLM types, $p>0.05$ and outperforms \textbf{No Active Perception} by an average of $6\%$, $p>0.05$. Using an Oracle VLM allows \textbf{Ours-LLM} to close the gap with the \textbf{Oracle} by an average of $5\%$ more than using InstructBLIP.
%Notably, using the Oracle VLM, our approach significantly outperforms \textbf{No Question} by an average of $30.5\%$, $p<0.01$, outperforms \textbf{No Active Perception} by an average of $6\%$, $p=0.1$, and outperforms \textbf{Baseline Questions} by an average of $4\%$, $p=0.3$. We under perform \textbf{Oracle} by $6\%$, $p=0.1$. 
Although our approach outperforms baselines using both VLMs, we suspect that InstructBLIP gives lower accuracies because the \dataset{} images -- especially the close-up images -- are out of distribution. For this reason, we presume that our approach gives a smaller advantage over other baseline methods when using InstructBLIP.

These results suggest that asking questions and actively perceiving the environment can enable grounded commonsense reasoning; with better VLMs, we can reach close to human-level performance. However, we were puzzled why the human \textbf{Oracle} was not more accurate. We hypothesize that in some situations, it is unclear what the most appropriate way to clean an object would be -- 
our annotators agreed $67\%$ of the time. To obtain higher accuracy, commonsense reasoning may sometimes not be enough and we must query user preferences to personalize the cleaning action; we explore this further in \autoref{sec:discussion} and the supplementary. We now analyze each component of our framework.

\begin{figure*}
    \centering
    \includegraphics[scale=0.4]{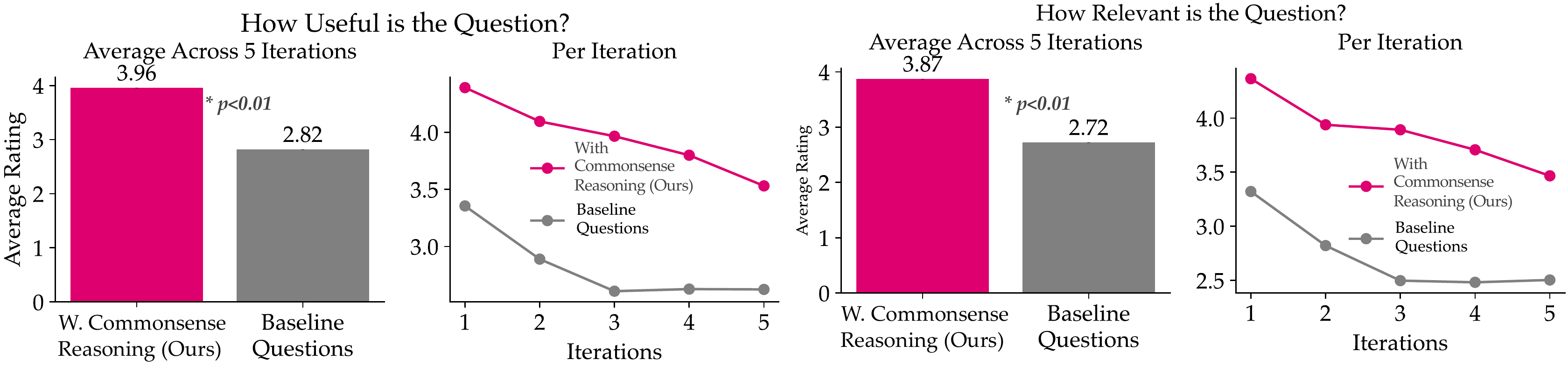}
    \caption{\small \textbf{How Good are the Follow-Up Questions?} Users rated our questions to be significantly more useful and relevant compared to baseline questions, $p<0.01$. However, the average usefulness and relevance of questions decreased over iterations.}
    \label{fig:question}
\end{figure*}

\noindent \textbf{Does the LLM Ask Good Follow-Up Questions?}
We first evaluate the LLM's follow-up questions and the reasoning used to produce those questions. On average, $82\%$ of users agreed that the reasoning was valid and $87\%$ agreed that the reasoning was appropriate. To evaluate the follow-up questions, we asked users to rate each question's usefulness and relevance for tidying the surface on a $5$-point Likert scale. We compared against \textbf{Baseline Questions}, where we removed the constraint that LLM-generated questions must be relevant for commonsense reasoning about normative values. An example baseline question is, ``Does the cup have a logo?'' All prompts and example questions are in the supplementary. \textbf{Users rated our questions to be significantly more useful and relevant for tidying surfaces compared to the baseline} ($p<0.01$, \autoref{fig:question}). However, across iterations, the average usefulness and relevance of our questions decreased. This result may be because there are not many useful and relevant questions to ask about simple objects such as a keyboard without interacting with them or people in the room.

% \begin{figure}
%     \centering
%     \adjincludegraphics[width=\textwidth,trim={0 0 {0.5\width} 0},clip]{figures/question.pdf}
% 	\adjincludegraphics[width=\textwidth,trim={{0.5\width} 0 0 0},clip]{figures/question.pdf}
%     \caption{\textbf{Evaluate Questions}. Ill make em larger}
%     \label{fig:question}
% \end{figure}

% \begin{figure*}
%     \centering
%     \includegraphics[width=\textwidth]{figures/question.pdf}
%     \caption{\small \textbf{How Good are the Follow-Up Questions?} Users rated our questions to be significantly more useful and relevant compared to baseline questions, $p<0.01$. However, the average usefulness and relevance of questions decreased over iterations.}
%     \label{fig:question}
% \end{figure*}

\begin{figure}
    \centering
    \includegraphics[width=\linewidth]{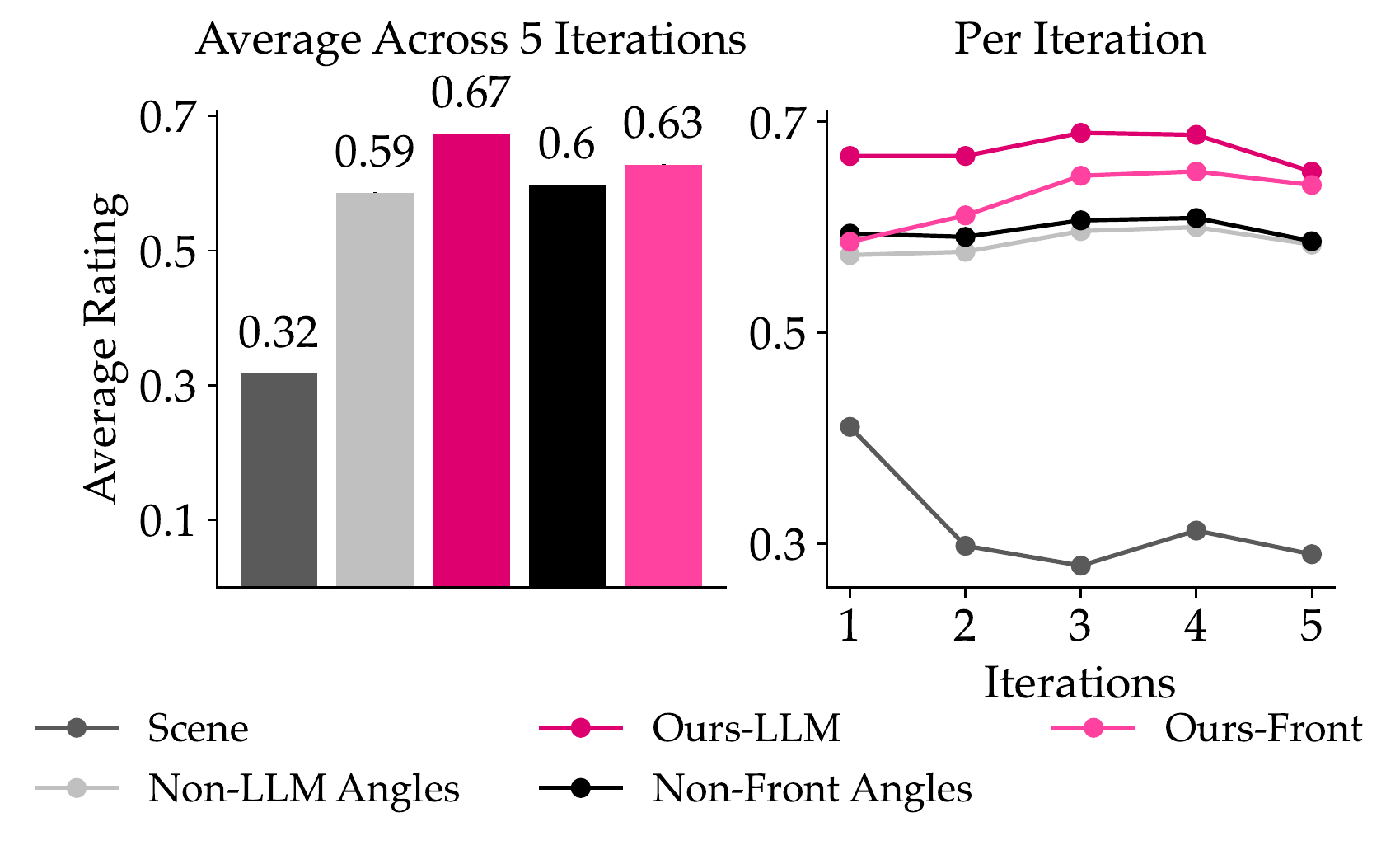}
    \caption{\small \textbf{Do We Choose Informative Close-Up Angles?} An average of $33.25\%$ more questions are answerable by the LLM-chosen angles and front angles compared to the scene, $p<0.01$. The LLM-chosen angles and front angle are also significantly more informative than the non-LLM-chosen angles and non-front angles respectively.}
    \label{fig:close_up}
\end{figure}

\noindent \textbf{Does the LLM Suggest Informative Close-Up Angles?} We next focus on whether the close-up angles suggested by the LLM are informative. For each object, we asked users whether the object's follow-up question is answerable from the close-up angle chosen by the LLM by showing them the corresponding close-up image. We also do this for the ``front'' angle. As our main baseline, we ask users whether questions are answerable from the scene-level view. Additionally, we compare against angles that the LLM did not choose (``Non-LLM Angles"), as well as non-front angles. \textbf{Across $5$ iterations we find that, on average, $35.5\%$ more questions are answerable by LLM-chosen angles and $31\%$ more questions are answerable by the front angles compared to the scene, $p<0.01$. The LLM-chosen angles and front angle are also significantly more informative than the non-LLM-chosen angles and non-front angles respectively.} This trend holds for each iteration (\autoref{fig:close_up}). 

\begin{table*}[h]
    \centering
    \caption{\small VLM multiple-choice prediction accuracy (zero-shot) under different angles over 5 iterations}
    \begin{tabular}{l c c c c c}
    \toprule
         & Scene & Non-front Angles & Front Angle & Non-LLM Angles &  LLM Angle  \\
     \midrule
      InstructBLIP (Vicuna) & 47.98 & 51.06 &  \textbf{52.64} & 50.94 & \textbf{53.21} \\
      InstructBLIP (Flan-T5) & 51.95 & 53.99 &  \textbf{56.74} & 54.08 & \textbf{56.30}  \\
     \bottomrule
    \end{tabular}
    \label{tab:vlm-perf}
\end{table*}

\noindent \textbf{Do Our Close-Up Angles Improve VLM Accuracy?}
Using VLMs for grounded commonsense reasoning pose challenges when there are obstructions in the image (e.g., a bag blocking the sports car) or when they are not able to describe relevant details. We hypothesized that providing a close-up image would ``help'' a VLM answer follow-up questions more accurately. We evaluate whether close-up images can actually improve VLM accuracy on follow-up questions. From the results in \autoref{tab:vlm-perf}, we see that \textbf{having access to close-up angles greatly improves the zero-shot prediction accuracy for both VLM variants.} More importantly, the front angles and the LLM proposed angles generally outperform other angles. These results show that it is beneficial to have both active perception and correct angles for our tasks.

\subsection{Evaluation on Real-World Robotic System}

% \begin{figure*}[t]
%     \centering
%     \includegraphics[width=\textwidth]{figures/robot_pipeline.pdf}
%     \caption{\small \textbf{Real-World Commonsense Reasoning.} We outline the steps of our framework with a robot. Notably, the LLM generates questions and ``angles'' for the arm to servo to (e.g., \textit{right of the banana}). We also use the LLM to generate an \textit{action plan} for each object -- each plan is converted to a sequence of skill primitives that are then executed by the robot.}
%     \label{fig:robot-pipeline}
%     \vspace*{-5mm}
% \end{figure*}

To assess the performance of our system on a real-world robot, we assemble $2$ surfaces with $11$ objects that require complex commonsense reasoning to tidy up. Importantly, we design these surfaces so that the commonsense way to tidy each object would be unambiguous. The first surface resembles a child's play area, with toys of ranging complexities (e.g., a MEGA Bloks structure, a partially built toy train set, and a to-scale Lego model of an Italian sports car). The robot must understand which toys to clean up and which toys should be left on display. The second surface, shown in \autoref{appx:robot-evaluation}, consists of trash that a robot must sort through and decide whether to recycle, put in landfill, or keep on the desk.

\noindent \textbf{Grounding Language in Robot Behavior.} Following the active perception component of our framework, we use a robot arm (equipped with a wrist camera) to servo to angles produced by the LLM and take photos. To map the LLM-produced angles and natural-language action plans to robot behavior, we implement a series of programmatic skill primitives (e.g., \texttt{relocate(``block'')}). In this work, each ``view'' and ``action'' primitive is defined assuming access to the ground-truth object class and position. These programmatic skill primitives define an API that the LLM can call into, similar to the process introduced by \citet{liangCodePoliciesLanguage2023}. Each action plan is translated to a sequence of these programmatic skills, which are then executed in an open loop (further implementation details are in the supplementary).

\noindent \textbf{Benchmark Evaluation Results.} To evaluate our method, we designed benchmark questions for each of the $11$ objects in a similar manner to that outlined in \autoref{sec:dataset}. We recruited $5$ annotators on Prolific to choose the correct answer and took the majority label. We report results for both the Oracle VLM and InstructBLIP after running $5$ iterations of our framework (see Figure in the supplementary). \textbf{Across both types of VLMs, \textbf{Ours-LLM} beats \textbf{Baseline Questions} by an average of $13.5\%$, beats \textbf{No Active Perception} by an average of $18\%$, and beats \textbf{No Questions} by an average of $13.5\%$.} With the Oracle VLM, we achieve \textbf{Oracle} performance. With InstructBLIP, our method produces a smaller advantage over baselines.

\section{Discussion}
\label{sec:discussion}
The purpose of this work is to equip robots with basic grounded commonsense reasoning skills to reduce the need for human specification. These reasoning skills can later be personalized towards an individual's preferences. To this end, we conduct a preliminary study to explore how we can add personalization on top of our framework. We analyzed questions that the human \textbf{Oracle} got incorrect in \autoref{sec:experiments} and found that object attributes such as ``dirtiness'' can indeed be subjective. This may have caused the \textbf{Oracle} to incorrectly answer some questions. We experimented with adding personalization information to $8$ questions where both the \textbf{Oracle} and our framework chose the same incorrect answer. \textbf{We found an average $86\%$ improvement in accuracy, suggesting that preference information helps further enable grounded commonsense reasoning.} See the supplementary for more details.

\noindent \textbf{Limitations and Future Work.} While our work presents a first step towards actively grounded commonsense reasoning, there are some limitations to address. One limitation is our reliance on heuristics to guide our active perception pipeline -- while the five specified angles are enough for most of the questions in the \dataset{} dataset, there are cases where objects may be occluded, or otherwise require more granular views to answer questions; future work might explore learned approaches for guiding perception based on uncertainty, or developing multi-view, queryable scene representations \citep{mildenhall2021nerf, kerrLERFLanguageEmbedded2023}. Similarly, we are limited by an inability to \textit{interact with objects dynamically} -- e.g., opening boxes, removing clutter. Finally, while we focus on commonsense behaviors, there are times where the ``right'' thing to do is to ask for human preferences.

\noindent \textbf{Acknowledgements.} This work was supported by DARPA YFA, NSF Award \#2006388, \#2125511, \#2218760, AFOSR YIP, JP Morgan, ONR, and TRI. 
\newpage
%This work makes a first step towards building competent robot assistants that reduce the need for human specification of commonsense behavior. We hope that future work can build on our framework and grow the various types of reasoning we want of our robots, enabling richer modes of human-robot interaction.

\bibliographystyle{IEEEtran}
\bibliography{IEEEabrv,references}

\clearpage
\appendix

%\section*{Supplementary Material}

% === Individual Appendix Sections ===

\subsection{Data Collection}
\label{appx:data-collection}
Our data collection consists of three components: 
\begin{enumerate}
    \item Collecting the \dataset{} dataset photos.
    \item Asking crowdworkers to choose the most appropriate action in our benchmark questions.
    \item Asking crowdworkers to evaluate parts of our framework.
\end{enumerate}

\subsubsection{Survey Interface}

We show the survey interface we used to complete the 2nd and 3rd crowdsourcing components below: 

\begin{figure*}[h!]
    \centering
    \includegraphics[width=\textwidth]{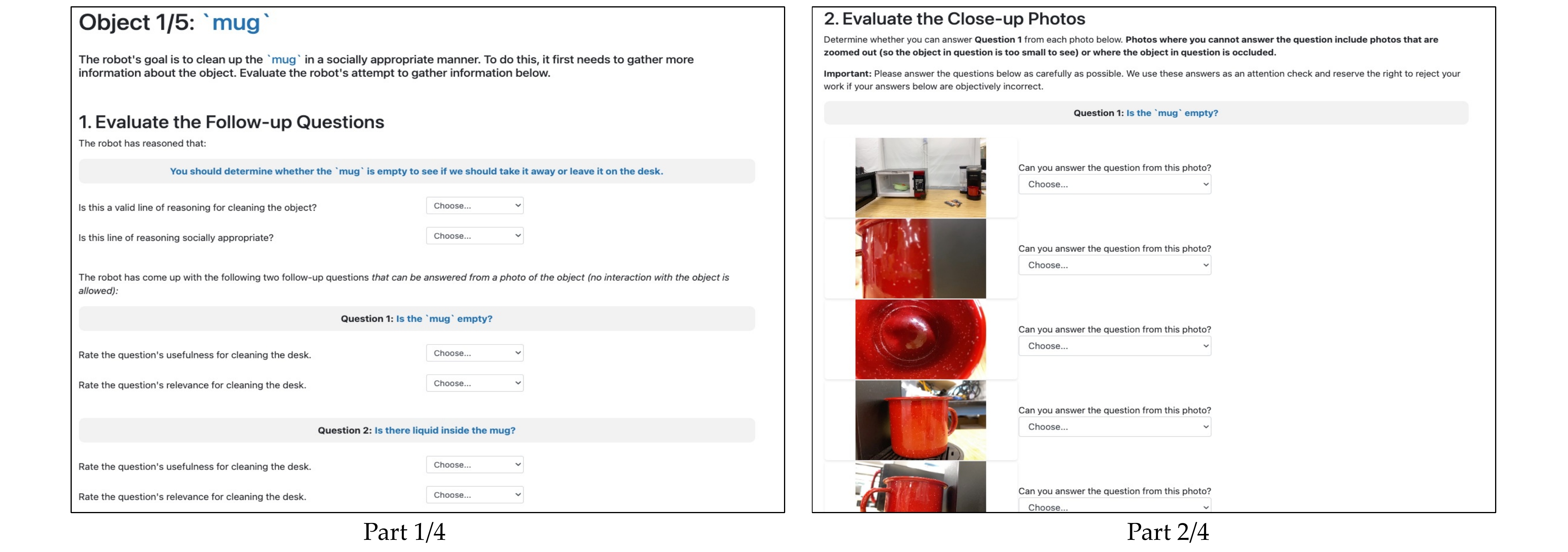}
    \caption{\textbf{Parts 1 and 2 of Survey Interface.} }
    \label{fig:survey1}
\end{figure*}

\begin{figure*}[h!]
    \centering
    \includegraphics[width=\textwidth]{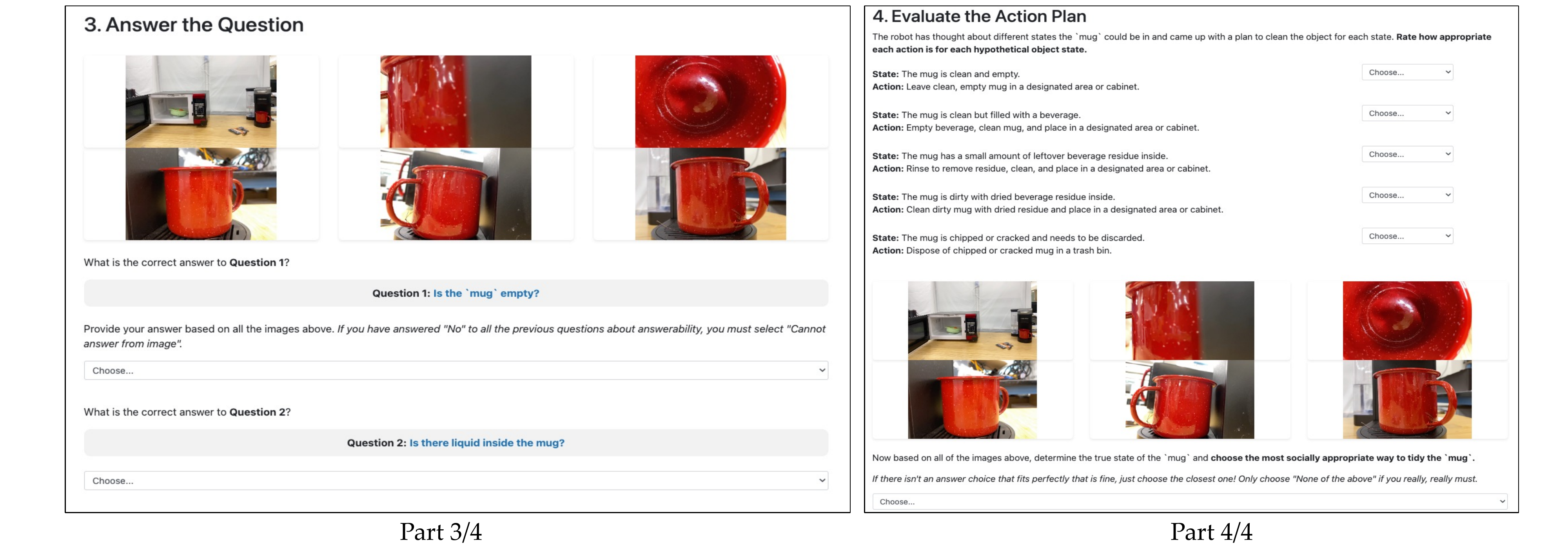}
    \caption{\textbf{Parts 3 and 4 of Survey Interface.} }
    \label{fig:survey2}
\end{figure*}

The survey consists of a set of questions that we ask about each object, with a single page per object. An example page for the ``mug'' object is shown in \autoref{fig:survey1} and \autoref{fig:survey2}. The first part of the survey asks users to rate the follow-up questions generated by the LLM; results are reported in Section 5 -- Experiments in main body of the paper, under \textit{``Does the LLM Ask Good Follow-Up Questions?''} The second part of the survey asks users to evaluate the informativeness of each close-up angle; results are also reported in Section 5, under \textit{``Does the LLM Suggest Informative Close-Up Angles?''} The third part of the survey asks users to give ground truth answers to the follow-up questions based on all six images collected of the object; these answers are used as the Oracle VLM when evaluating our framework. The final part of the survey asks users to evaluate the appropriateness of each multiple choice option in the \dataset{} benchmark and asks them to indicate the most appropriate way to tidy the object. These results are used to determine the correct answer for our benchmark questions as described in Section 4 of the main paper. We designed our survey using Flask and Python and hosted it on an AWS server.

\subsubsection{Prolific Details}

We recruited crowdworkers from Prolific to complete our study. The study took an average of $10$ minutes and each crowdworker was paid \$$2$ (\$$12$/hour). We required workers to be fluent English speakers, based in the U.S. or U.K., and have a minimum approval rating of $98\%$. Each worker was in charge of answering survey questions about all objects belonging to a desk. We have a total of $70$ desks and ran our framework $5$ times, resulting in the recruitment of $350$ Prolific workers.

\subsection{Framework Implementation Details}
\label{appx:evaluation-details}
In this section, we describe the design choices and implementation details of our framework. 

\noindent \textbf{Generating the Initial Description.} In the first step of our framework (Section 3 in the main paper), we generate an initial description of the scene and append it to our context $\mathcal{C}^0$. The initial description is a list of all the objects in the scene. To ensure that we list the objects accurately, we generate the initial description using ground truth names of objects (see \autoref{lst:initial_description} for an example).

\begin{lstlisting}[language=Python, caption=Example Initial Description, label=lst:initial_description]
"""
These are the objects on the desk: 
    `scrunchie`, `lotion`, `vaseline`, `brush`.
"""
\end{lstlisting}

% \begin{figure*}
%     \centering
%     \includegraphics[width=\textwidth]{figures/question.pdf}
%     \caption{\small \textbf{How Good are the Follow-Up Questions?} Users rated our questions to be significantly more useful and relevant compared to baseline questions, $p<0.01$. However, the average usefulness and relevance of questions decreased over iterations.}
%     \label{fig:question}
% \end{figure*}

\noindent \textbf{Structuring Follow-Up Questions.} In the second step of our framework, we prompt an LLM to generate follow-up questions about information that it is missing in its context. We structure our follow-up questions to be yes-or-no questions where the LLM also has an option to choose ``Cannot answer from image''. We choose a yes-or-no question format to make it easier to evaluate the VLM's answers to these question. See \autoref{appx-subsec:prompt_question} for the actual prompts used for the LLM.

\noindent \textbf{Eliciting Informative Close-Up Angles from an LLM.} In the third step of our framework, we prompt an LLM to generate informative close-up angles that guide a photo-taking robot. We restrict the close-up angles the LLM can choose to a set of $5$ angles: \texttt{<FRONT>, <BACK>, <LEFT>, <RIGHT>, <TOP>}. When querying the LLM, we format the prompt as a multiple choice question where the options are the five specified angles. See \autoref{appx-subsec:prompt_question} for further prompting details. 

\subsection{Real-World Robot Evaluation}
\label{appx:robot-evaluation}
% === Real-World Robot Evaluation ===
\begin{figure*}[t]
    \centering
    \includegraphics[width=\textwidth]{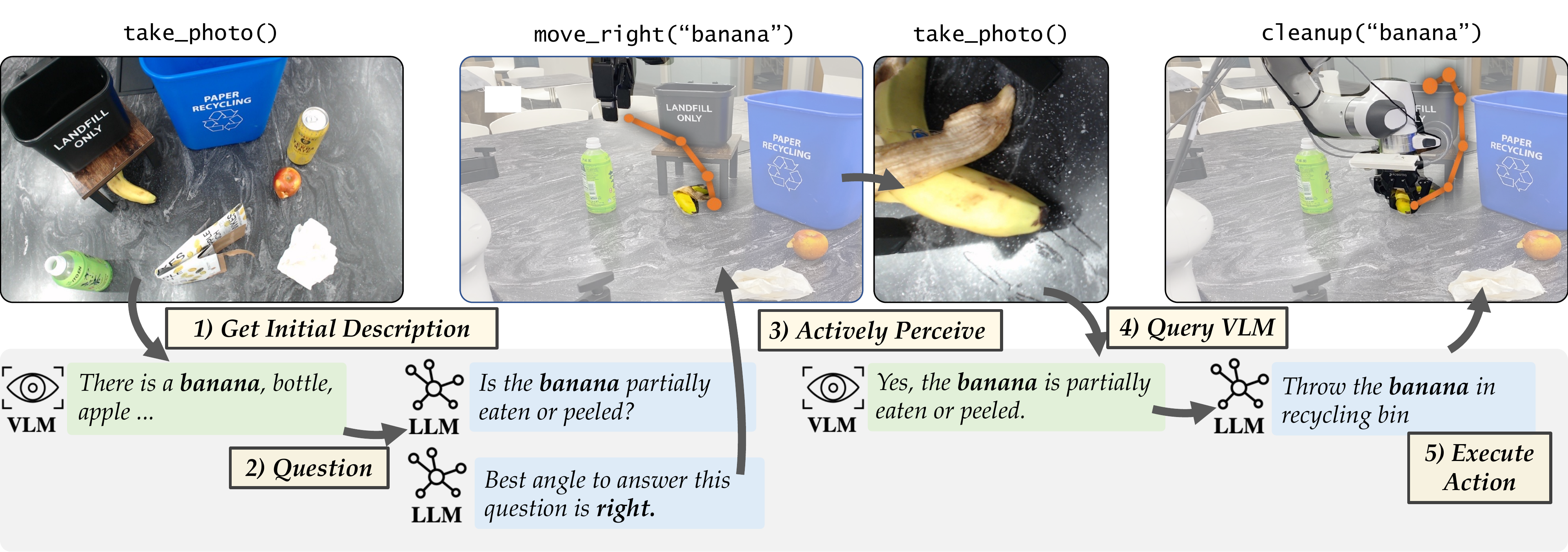}
    \caption{\small \textbf{Real-World Commonsense Reasoning.} We outline the steps of our framework with a robot. Notably, the LLM generates questions and ``angles'' for the arm to servo to (e.g., \textit{right of the banana}). We also use the LLM to generate an \textit{action plan} for each object -- each plan is converted to a sequence of skill primitives that are then executed by the robot.}
    \label{fig:robot-pipeline}
\end{figure*}
% === Body Text ===
When implementing our grounded commonsense reasoning system on physical robot hardware (\autoref{fig:robot-pipeline}), there are two operating modes, reflecting the \textit{active perception} and \textit{skill execution} components of our approach respectively. As a preliminary, for the real-robot experiments, we assume that the object poses (in the coordinate frame of the robot's end-effector) are known a priori. While in this work we assume these poses are hand-specified by an expert, one could also use off-the-shelf perception systems that predict 6-DoF object poses or bounding boxes directly, as in prior work \citep{wuTidyBotPersonalizedRobot2023}.

\noindent \textbf{Active Perception Primitives.} The active perception component of our framework requires the robot to execute on two types of behaviors, which we codify as functional primitives \texttt{move\_<direction>(<object>)} and \texttt{take\_photo()}. While the latter behavior is well-defined (capture an image at the robot's current position), the directional movement primitives vary \textit{per-object}. As each object in our experiments is of different scale and composition, we manually define a set of pose transformations $p_\text{dir} \in \text{SE}(3)$ for each object and direction \texttt{<FRONT>, <BACK>, <LEFT>, <RIGHT>, <TOP>}. Given this dictionary of pose offsets, we implement \texttt{move\_direction} for a specified object and desired direction by planning and executing a min-jerk trajectory from the robot's current location to the resulting pose after applying $p_\text{dir}$ to the known object's pose.

\begin{figure*}[t]
    \centering
    \includegraphics[scale=0.15]{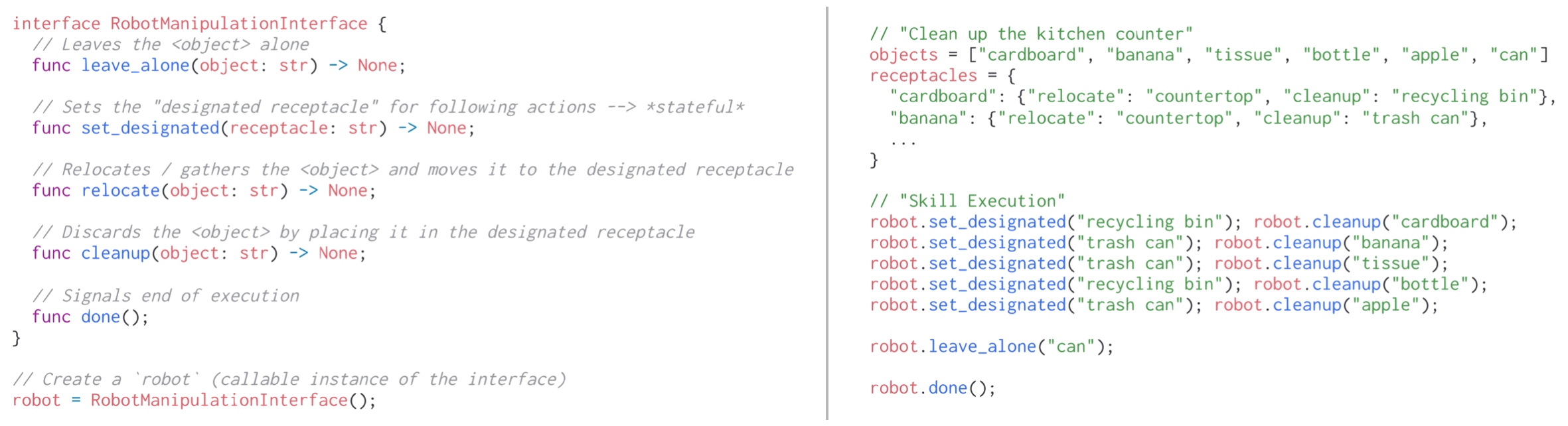}
    \caption{\textbf{Code as Policies Interface for Real-Robot Execution.} We define a simple programmatic interface for specifying robot skill primitives on in an \textit{object-oriented fashion}. The interface is stateful; for robot primitives such as \texttt{cleanup()} and \texttt{relocate()}, the robot sets a designated receptacle via the special function \texttt{set\_designated()}. On the right, we provide the actual execution trace produced by the LLM for the Kitchen Cleanup Desk (see \autoref{fig:kitchen_cleanup_desk}).}
    \label{fig:cap-interface}
\end{figure*}

\begin{figure*}
    \centering
    \includegraphics[width=.95\textwidth]{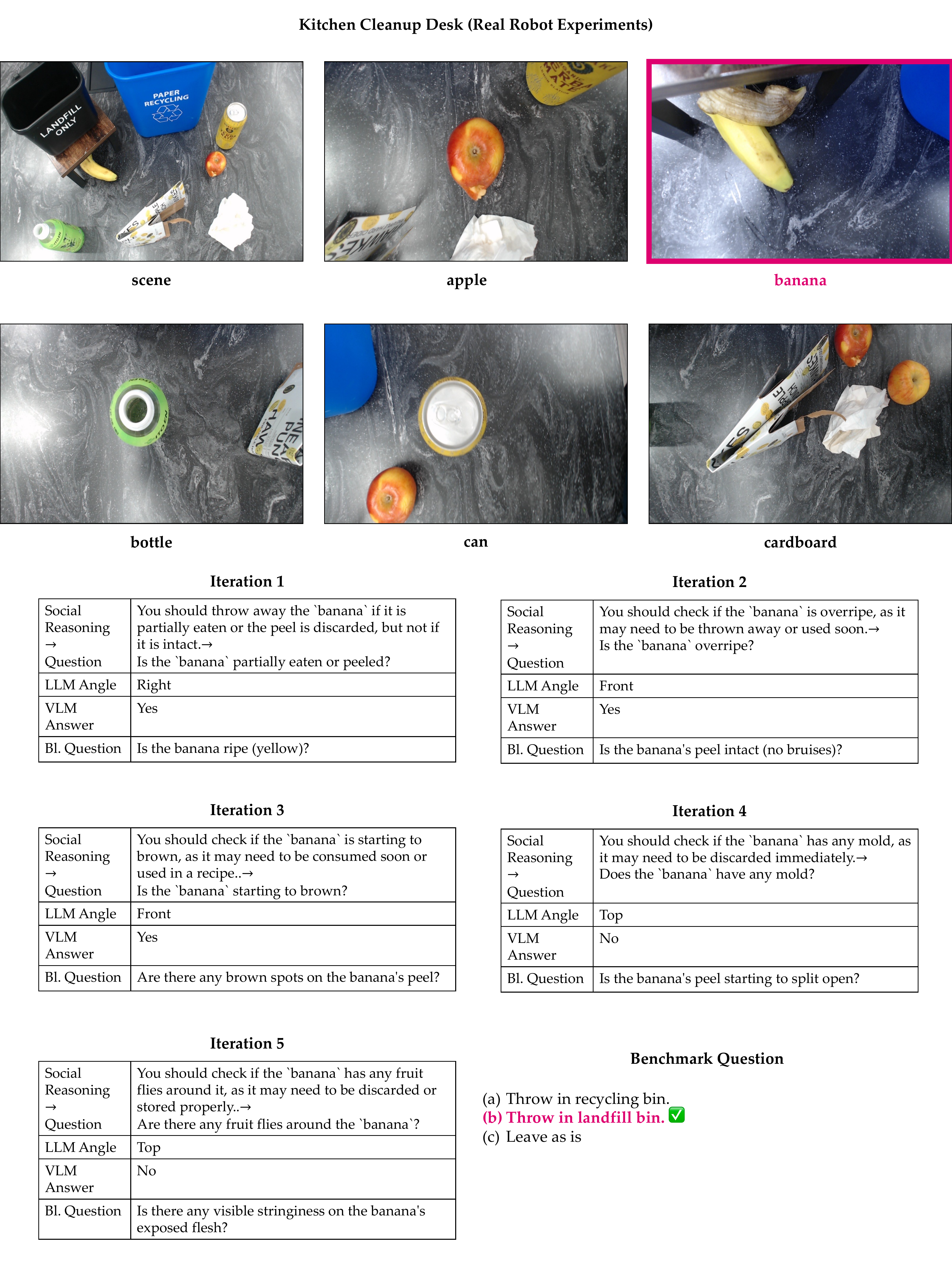}
    \caption{\textbf{Kitchen Cleanup Desk (Real-Robot Experiments).} Example photos of each object (from a top-down angle). We also provide examples of our framework's outputs for the \emph{banana} for all $5$ iterations using InstructBLIP.}
    \label{fig:kitchen_cleanup_desk}
\end{figure*}

\noindent \textbf{Implementing Object-Centric Manipulation Skills.} Similar to the perception primitives, we define each manipulation skill on a per-object basis as well; this is both due to the variety in object scale and properties, but also due to the variance in grasp locations for different desired behaviors. For example, the location where the robot should grasp an object such as a soda can may differ greatly depending on whether we want to throw the soda can away into a recycling bin (in which case the robot should grasp the soda can across the top), or if we want to relocate the can to a shelf (in which case the robot should grasp the soda can along the side, to aid in insertion). To formalize this, we define a fixed interface depicted in \autoref{fig:cap-interface}. The provided API defines functions for each skill -- for example, \texttt{relocate()} and \texttt{cleanup()} -- at the \textit{object-level}, with a stateful function \texttt{set\_designated()} that provides a compositional way to set target locations (i.e., ``receptacles''). \autoref{fig:cap-interface} (Right) shows the actual invoked API calls for the Kitchen Cleanup Desk depicted in \autoref{fig:kitchen_cleanup_desk}.

We implement each object-oriented skill -- \texttt{relocate()} and \texttt{cleanup()} -- for a given object $o_i$ and receptacle $r_j$ as a tuple of pick-and-place waypoints defined as ($\text{pick}_{o_i} \in \text{SE}(3)$, $\text{place}_{r_j} \in \text{SE}(3)$; each pick/place point is defined as a transformation relative to the origin of the robot's reference frame. To execute on a ``pick'' waypoint, we plan a collision-free min-jerk trajectory to the given pose, and execute a blocking grasp action; similarly, to execute on a ``place'' waypoint, we plan a similar trajectory to the given receptacle pose, and execute a blocking open-gripper action. We run all experiments with a 7-DoF Franka Emika Panda robot manipulator equipped with a Robotiq 2F-85 gripper, using Polymetis \citep{polymetis} to facilitate real-time control and Pinocchio \citep{carpentier2019pinocchio, pinocchioweb} for trajectory planning.

\noindent \textbf{Grounding Language to Skills.} While the API definition deterministically defines robot behavior and skill execution in a given environment, we need a way of mapping natural langauge action plans generated by the LLM to sequence of API calls -- for example, mapping the language action ``dispose of the coffee cup'' to the corresponding API calls \texttt{robot.set\_designated(``recycling bin''); robot.cleanup(``coffee cup''); robot.done()}. To do this, we follow a similar procedure as in prior work using LLMs for code generation, prompting an LLM with the API definition, a series of in-context examples, and a continuation prompt for generating a valid sequence of API calls. The continuation prompt contains the set of known objects in the environment and valid receptacles defined for each skill, following prior work \citep{ahnCanNotSay2022,liangCodePoliciesLanguage2023}. The full prompt is in \autoref{appx-subsec:cap-prompts}.

\begin{figure*}
    \centering
    \includegraphics[width=\textwidth]{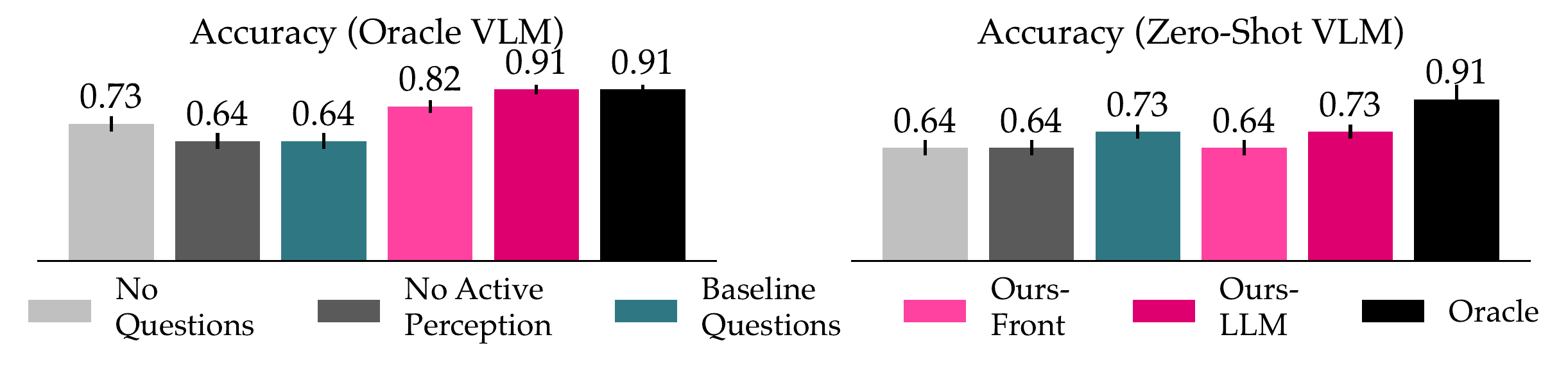}
    \caption{\small \textbf{Real Robot Benchmark Accuracy}. We construct benchmark questions for objects used with the real robot in similar manner to Section 4 in the main paper. Across both types of VLMs, our \textbf{Ours-LLM} beats \textbf{Baseline Questions} by an average of $13.5\%$, beats \textbf{No Active Perception} by an average of $18\%$, and beats \textbf{No Questions} by an average of $13.5\%$.}
    \label{fig:robot-benchmark}
\end{figure*}

\noindent \textbf{Evaluation.} We add \autoref{fig:robot-benchmark} to supplement our results described in Section 5 of the main paper.

% === Code as Policies Interface & Execution Trace for Food Desk ===

\subsection{VLM Details}
\label{appx:vlm-details}
\begin{figure*}[t]
    \centering
    \begin{minipage}{.7\linewidth}
        \centering
\begin{lstlisting}[language=Python]
"""
Given the image, please answer the following 
    question in yes, no, or unknown.
Question: Is the bagel sandwich partially eaten?
Answer: 
"""
\end{lstlisting}
    \end{minipage}
    \begin{minipage}{0.25\linewidth}
        \centering
        \includegraphics[width=0.6\linewidth]{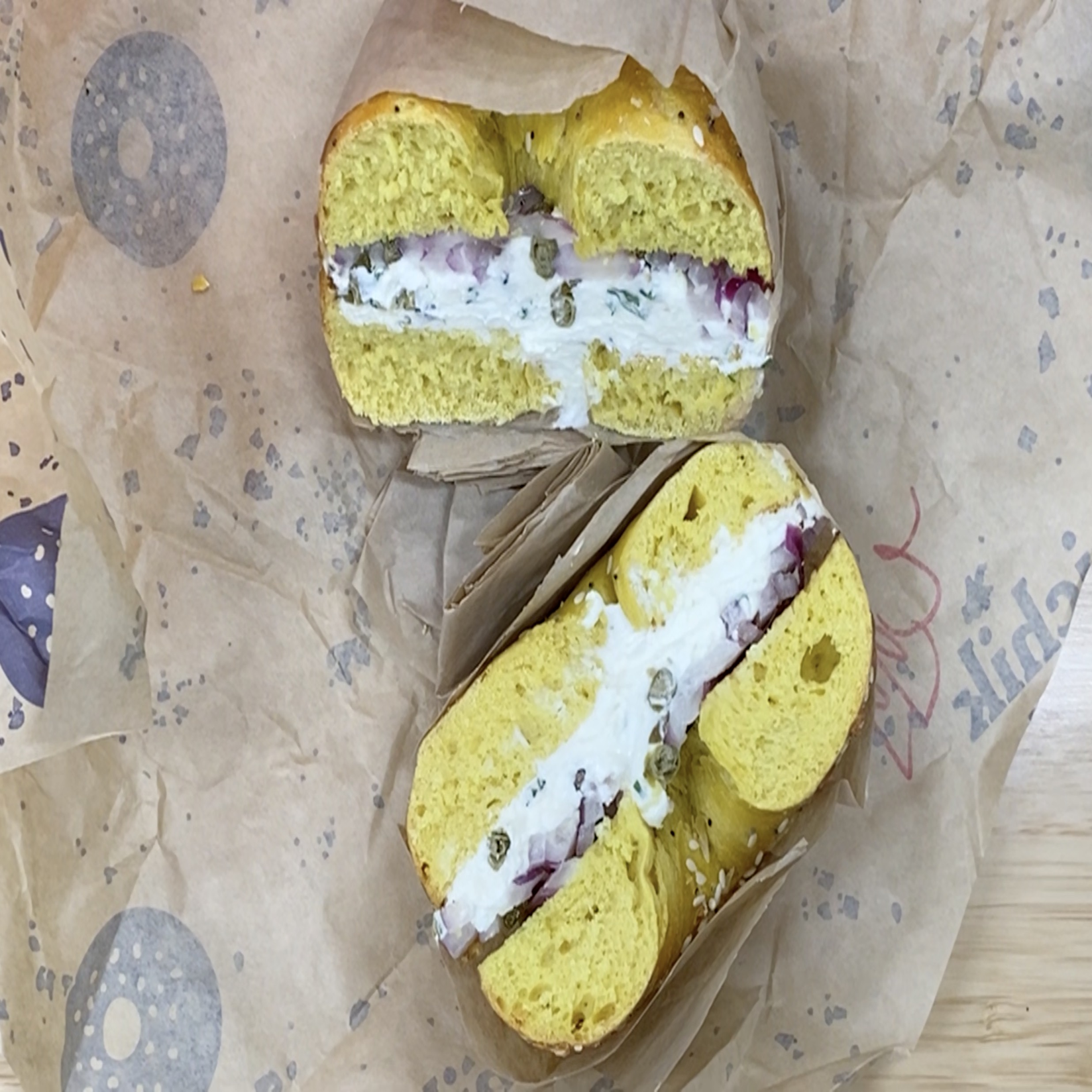}
    \end{minipage}
    \caption{Example of VLM Text Prompt and Image Input.}
    \label{fig:vlm-prompt}
\end{figure*}

We use pretrained visual-and-language models (VLMs) trained on massive internet scale images and texts to answer the questions generated by LLM. Following \autoref{appx:evaluation-details}, we prompt the LLM so that it generates queries that can be easily answered by \emph{yes}, \emph{no} or \emph{unknown}; these queries (and the respective images) are the inputs to the VLM.

To make it easier to parse the predictions of the VLM question-answerer, we rank the three answer options conditioned on the image and text input, rather than allowing the VLM to generate free-form responses. Specifically, we set the text prompt following \autoref{fig:vlm-prompt}. We use InstructBLIP \citep{dai2023instructblip} as our VLM and select the output with the highest predicted probability $P(\texttt{answer} \mid \texttt{prompt}, \texttt{image})$ for \texttt{answer} $\in$ ~\{\emph{yes}, \emph{no}, \emph{unknown}\} as the final answer. As InstructBLIP can use multiple LLM backbones, we evaluate both the Vicuna-13B and Flan-T5-XXL (11B) variants, finding Flan-T5-XXL to work better for our tasks. We have also experimented with further finetuning InstructBLIP on the in-domain data from the \dataset{} dataset, but have not seen any noticeable performance gains; as a result, we use the off-the-shelf pretrained models in this work.

\subsection{Personalization Analysis}
\label{appx:personalization}
\begin{figure*}
    \centering
    \includegraphics[width=.95\textwidth]{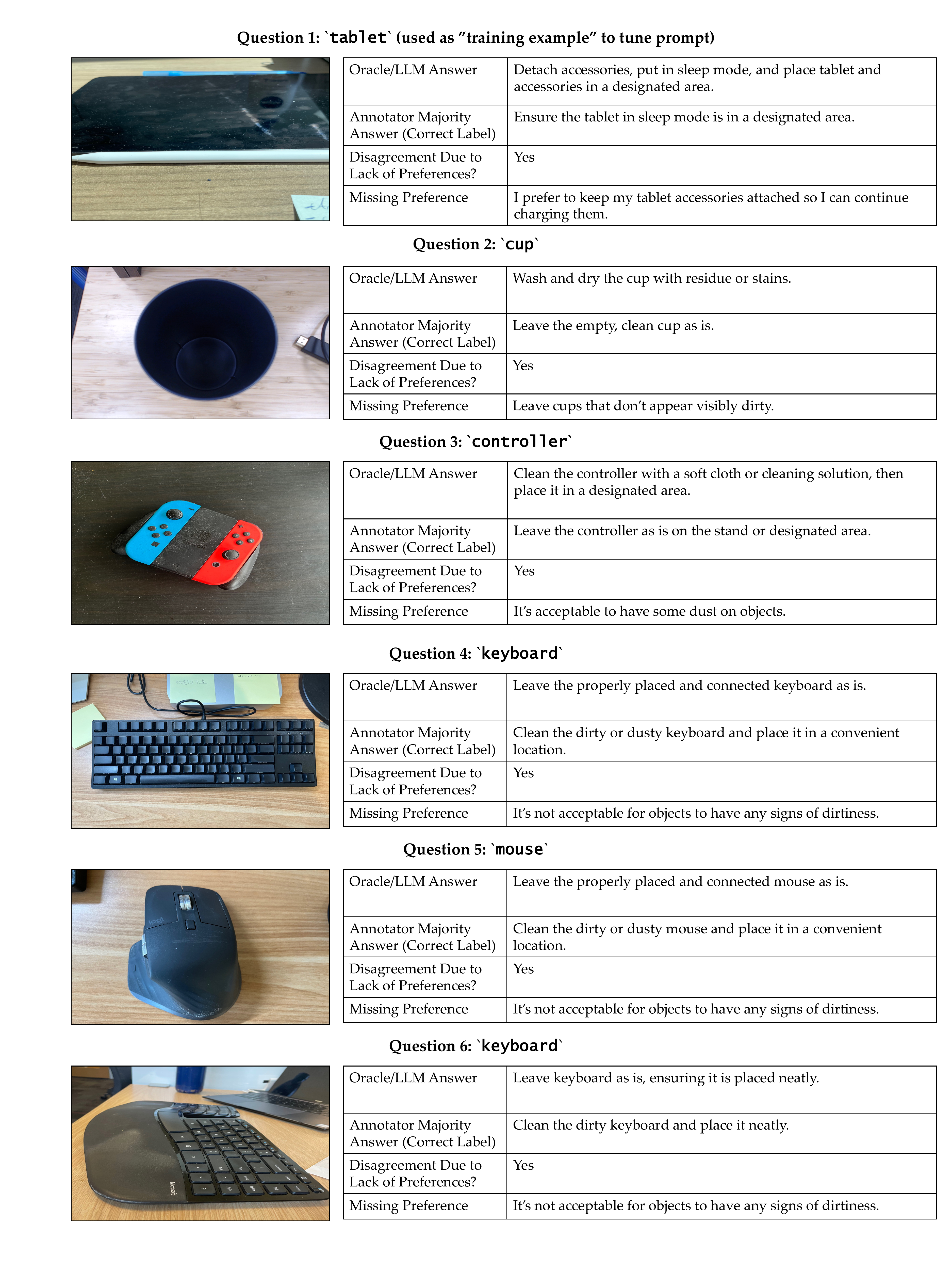}
    \caption{\textbf{Questions Used for Personalization Analysis (1/2).} We display questions where both \textbf{Oracle} and \textbf{Ours-LLM} chose the same incorrect answer. We recruited a participant to indicate whether the \textbf{Oracle} could have incorrectly answered these questions due to a lack of preference information, and if so, what the preference would be.}
    \label{fig:personalization1}
\end{figure*}

We explore the hypothesis that incorporating personal preferences on how to clean objects can lead to a higher accuracy on our benchmark, as discussed in Sections $5$ and $6$ of the main paper. We studied questions that the human \textbf{Oracle} got incorrect in Section $5$ of the main paper. Qualitatively, we found that some attributes of an object such as its ``dirtiness'' can be subjective, lending support to our hypothesis. This may have caused the \textbf{Oracle} to incorrectly answer some questions. For instance, in Question $6$ of \autoref{fig:personalization1}, the \textbf{Oracle} did not consider a keyboard that had a small amount of dust on it to be ``dirty'' enough and chose to ``leave it as is''. However, the majority of annotators preferred that the keyboard ``should be cleaned''.

\begin{figure*}
    \centering
    \includegraphics[width=.95\textwidth]{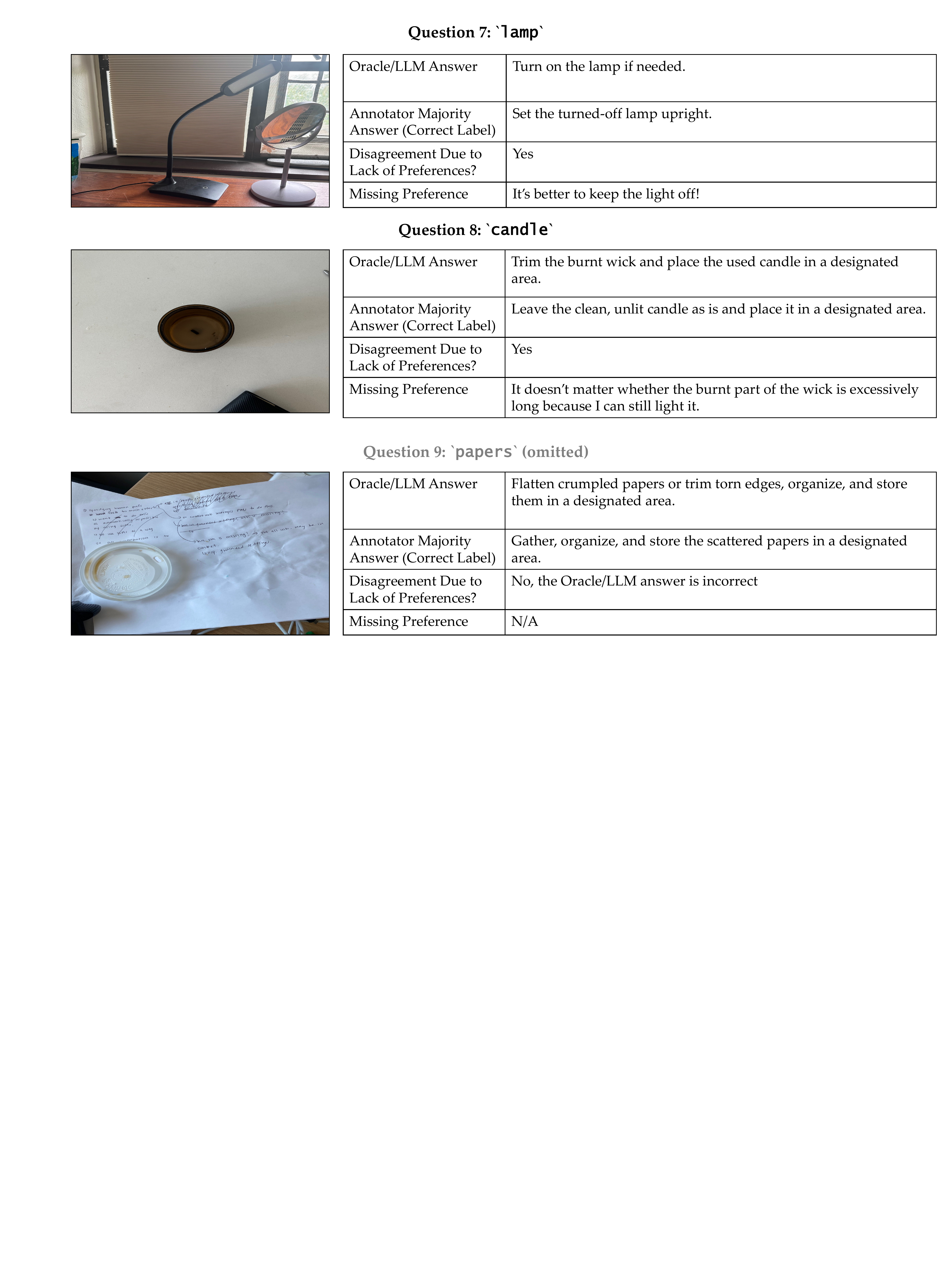}
    \caption{\textbf{Questions Used for Personalization Analysis (2/2).}  We display questions where both \textbf{Oracle} and \textbf{Ours-LLM} chose the same incorrect answer. We recruited a participant to indicate whether the \textbf{Oracle} could have incorrectly answered these questions due to a lack of preference information, and if so, what the preference would be.}
    \label{fig:personalization2}
\end{figure*}

We explored whether adding preferences would improve our framework's accuracy. We selected $9$ questions where both the \textbf{Oracle} and our framework, \textbf{LLM-Ours}, chose the same incorrect answer. The full list of questions is shown in \autoref{fig:personalization1} and \autoref{fig:personalization2}. We recruited a participant and, for each question, asked them whether the \textbf{Oracle} could have chosen the incorrect answer because of a lack of preference information. If the participant agreed that there was as lack of preference information, we asked them what the preference would be. For instance, in Question $6$ of \autoref{fig:personalization1}, the user noted that the disagreement between the human \textbf{Oracle} and human annotators could have been due to a lack of preference information, such as ``It's not acceptable for objects to have any signs of dirtiness''. The participant indicated that the \textbf{Oracle} could have incorrectly answered $8$ out of the $9$ questions due to a lack of preference information. Question $9$ in \autoref{fig:personalization2} is an example of a question where the user thought the \textbf{Oracle} was incorrect due to noise.

For the remaining $8$ questions, our goal was to see if adding preferences to the LLM's prompt would help the LLM choose the ``correct'' action plan as indicated by the annotators' majority label. We used $1$ question to tune the prompt and evaluated the LLM on the remaining $7$ questions (Questions $2-8$ in \autoref{fig:personalization1} and \autoref{fig:personalization2}). We prompted the LLM by appending preference information to the prompt for choosing an action plan (described in \autoref{appx-subsec:prompt-action-plan}). An example prompt is shown in \autoref{lst:personalization}:
\begin{lstlisting}[language=Python, caption=Example Prompt for Generation an Action Plan with Preference Information, label=lst:personalization]
"""
The owner of the object has a preference on how you should tidy the `candle`: Don't trim the wick. It doesn't matter whether the burnt part of the candle wick is excessively long because I can still light it.

The best option is:
"""
\end{lstlisting}

\textbf{We found an average $86\%$ improvement in accuracy, lending support to the hypothesis that preference information helps further enable grounded commonsense reasoning.}

\subsection{Prompts \& In-Context Examples for LLM Inference}
\label{appx:prompting-details}
In this section, we provide the comprehensive set of prompts used to elicit the desired behavior from the LLM (via the OpenAI API) across the multiple functionalities described in our approach, from generating follow-up questions, to synthesizing code for real-robot execution.

\subsubsection{Prompt for Generating Follow-Up Questions}
\label{appx-subsec:prompt_question}

In the second step of our proposed framework (see Section 3 of the main paper), we one-shot prompt the LLM to generate follow-up questions about a list of objects on a surface using the prompt in \autoref{lst:question}.

\begin{lstlisting}[language=Python, caption=Instruction For Generating Follow-Up Questions, label=lst:question]
"""
These are the objects on the desk: 
   `scrunchie`, `lotion`, `vaseline`, `brush`. 

Your goal is to tidy the desk in a socially appropriate manner. 
Ask a new follow-up question about each object to gather 
more information. Only ask questions that can be answered by 
taking a picture of the object. For example, DO NOT ask whether 
the object is currently being used.
"""
\end{lstlisting}

To guide follow-up question generation, we provide the following (\autoref{lst:question2}) as the sole in-context example before having the LLM generate a continuation:

\begin{lstlisting}[language=Python, caption=In-Context Example For Generating Follow-Up Questions, label=lst:question2]
"""
These are the objects on the desk: 
  `apple`, `charging cable`, `empty water bottle`, 
  `book`, `calendar`, `coffee cup`. 
  
Your goal is to tidy the desk in a socially appropriate manner. 
Ask a new follow-up question about each object to gather 
more information. Only ask questions that can be answered by 
taking a picture of the object. For example, DO NOT ask 
whether the object is currently being used.

-`Apple`: 
  Socially motivated reasoning: You should throw away the 
    `apple` if it is partially eaten, but not if it is intact.
  
  Resulting question (that can be answered by taking a 
    picture of object): Is the `apple` partially eaten? 
    
    (a) Yes (b) No (c) Cannot answer from image

-`Charging cable`: 
  Socially motivated reasoning: You should coil the 
    `charging cable` and store it neatly if it is not in use, 
    but leave it in place if it is connected to a device that 
    needs charging.

  Resulting question (that can be answered by taking a 
  picture of object): Is the `charging cable` connected to a device? 
  
  (a) Yes (b) No (c) Cannot answer from image

...
"""
\end{lstlisting}

Notably, we use Chain-of-Thought prompting to encourage the LLM to generate questions that are motivated by commonsense reasoning. We also encourage the LLM to ask questions that can be answered by an image of the object.

\noindent \textbf{Prompt for Generating Baseline Follow-Up Questions.}

To generate baseline questions, we use the following prompt (\autoref{lst:baseline_question}):

\begin{lstlisting}[language=Python, caption=Instruction For Generating Baseline Follow-Up Questions, label=lst:baseline_question]
"""
Ask one yes-or-no question for each object on the desk. Only ask 
yes-or-no questions that can be answered by taking a picture of the object.

These are the objects on the desk:
    `scrunchie`, `lotion`, `vaseline`, `brush`. 

Format your answer in the following format: `object_name`: question
"""
\end{lstlisting}

In our baseline question prompt, we do not specify that the goal for the LLM is to tidy the desk nor do we require the LLM to generate commonsense-motivated questions.

\subsubsection{Prompt for Choosing a Close-Up Angle}
\label{appx-subsec:closeup-angle}

In the third step of our proposed framework, we few-shot prompt the LLM to generate informative close-up angles that would guide a robot. In the prompt, we include a list of objects on the current surface, the follow-up question about an object, and a multiple choice question with options corresponding to the five predefined close-up angles: \texttt{<FRONT>, <BACK>, <LEFT>, <RIGHT>, <TOP>}. We use the following prompt (\autoref{lst:close_up_angle}):

\begin{lstlisting}[language=Python, caption=Prompt for Generating Informative Close-Up Angles, label=lst:close_up_angle]
"""
Description: These are the objects on the desk: 
    `computer monitor`, `cup`, `computer wires`, `apple`.

Follow-up Question: Are the `computer wires` connected to anything? 
    (a) Yes (b) No

You are instructing a robot to take a close-up picture of the object 
to help answer the follow-up question. 

Which of the following angles would yield a close-up picture that can 
best answer the question?

(a) Top of the object
(b) Right side of the object
(c) Left side of the object
(d) Front of the object
(e) Behind the object

Response: A top-down view would give an unoccluded view since the 
wires might be tangled. 

(a) Top of the object

Description: These are the objects on the desk: 
    `monitor`, `stack of papers`, `cups`.

Follow-up Question: Are the `cups` empty? 
    (a) Yes (b) No

You are instructing a robot to take a close-up picture of the object 
to help answer the follow-up question. 

Which of the following angles would yield a close-up picture that can 
best answer the question?

(a) Top of the object
(b) Right side of the object
(c) Left side of the object
(d) Front of the object
(e) Behind the object

Response: The cups might be opaque so the best angle would be 

(a) Top of the object

Description: These are the objects on the desk: 
    `keyboard`, `whiteboard marker`, `stack of papers`, `vase`.

Follow-up Question: Are the `stack of papers` straightened?
    (a) Yes (b) No

You are instructing a robot to take a close-up picture of the object 
to help answer the follow-up question. 

Which of the following angles would yield a close-up picture that can 
best answer the question?

(a) Top of the object
(b) Right side of the object
(c) Left side of the object
(d) Front of the object
(e) Behind the object

Response: The stack would best be viewed from its side. 

(d) Front of the object
"""
\end{lstlisting}

\subsubsection{Prompt for Choosing an Action Plan}
\label{appx-subsec:prompt-action-plan}

As the ultimate step of our framework, we prompt the LLM to answer our benchmark questions by choosing the most socially appropriate action to tidy each object. When prompting the LLM, we first include the context accumulated so far: the follow-up questions and their VLM-generated answers (see \autoref{lst:action_plan_answer} for an example).

\begin{lstlisting}[language=Python, caption=Example of Context for Action Plan Generation, label=lst:action_plan_answer]
"""
Here is some information about the `scrunchie` in 
question-answer format.

Is the `scrunchie` neatly placed on the desk? Yes
Does the `scrunchie` have any stains? Yes
Does the `scrunchie` have any loose threads? No
"""
\end{lstlisting}

We append the benchmark question to the prompt and have the LLM generate an appropriate tidying action:

\begin{lstlisting}[language=Python, caption=Prompt For Generating Answers to Benchmark Questions, label=lst:action_plan_answer2]
"""
Based on the information above, what is the most appropriate
way to tidy the `scrunchie`? 

Choose the best option.
(a) The scrunchie is neatly coiled and placed on the desk. 
  -> Leave the neatly coiled scrunchie as is in a designated area.
(b) The scrunchie is stretched out and tangled with other 
  items on the desk. 
  -> Untangle, coil neatly, and place in a designated area.
(c) The scrunchie is dirty or stained and needs to be cleaned. 
  -> Clean, dry, and place in a designated area.
(d) The scrunchie is partially unraveled or damaged. 
  -> Repair or replace, and place in a designated area.
(e) The scrunchie is being used to hold together a bundle 
  of cables or cords on the desk. 
  -> Remove from cables, coil neatly, and place in a designated area.

The best option is:
"""
\end{lstlisting}

\subsubsection{Prompt for Generating \dataset{} Benchmark Questions}
\label{appx-subsec:dataset-question-gen}

As described in Section 3 of the main paper, we prompt an LLM to generate multiple choice options for the question \textit{``What is the most appropriate way to tidy the object?''} for each object in the \dataset{} dataset. To generate each set of multiple choice options, we first prompt the LLM to list five possible states each object could be in:

\begin{lstlisting}[language=Python, caption=Example Prompt For Generating Benchmark Questions (1/2), label=lst:benchmark_q1]
"""
These are the objects on the desk: 
    `scrunchie`, `lotion`, `vaseline`, `brush`.

Your goal is to tidy each `object` up, but there is not 
enough information about each object. For each `object`, 
list 5 possible states the object could be in that would 
affect how you tidy it up. 

Label the 5 states (a)-(e). Make sure each state is 
significantly different from each other. Remember that 
all the objects are placed on the desk.
"""
\end{lstlisting}

After receiving the LLM's response, we ask it to generate a cleaning action for each state. The purpose of first asking it to generate object states is so that the LLM can generate diverse cleaning actions:

\begin{lstlisting}[language=Python, caption=Example Prompt For Generating Benchmark Questions (2/2), label=lst:benchmark_q2]
"""
For each state (a)-(e), tell me how you would tidy the `object`. 
Make sure each answer choice is significantly different from each 
other. Include an option to 'leave the object as is'. 
Each object should be in apostrophes like so: `object`.
"""
\end{lstlisting}

\subsubsection{Prompt for Real-Robot Code Generation from Language}
\label{appx-subsec:cap-prompts}

Following \autoref{appx:robot-evaluation}, we use the LLM to generate valid API calls for a given natural language action (e.g., ``dispose of the coffee cup''). To do this, we use the following instruction prompt for GPT-3 that defines the interface formally:

\begin{lstlisting}[language=Python, caption=Prompt for Generating Real-Robot API Calls from Natural Language Actions, label=lst:cap-prompt-api]
INITIAL_INSTRUCTION = (
    """
    Translate each of the following language instructions to a 
    sequence of predefined API calls that will be executed by 
    a robot manipulator to help "clean up" a workspace. 
    When generating code, make sure to use the API provided below:
    """
)

ROBOT_API = (
    """
    interface RobotManipulationInterface {
        // Leaves the <object> alone
        func leave_alone(object: str) -> None;
        
        // Sets the "designated receptacle" for 
        // following actions --> *stateful*
        func set_designated(receptacle: str) -> None;
        
        // Relocates / gathers the <object> and moves it to the 
        // designated receptacle
        func relocate(object: str) -> None;
    
        // Discards the <object> by placing it in the 
        // designated receptacle
        func cleanup(object: str) -> None;
    
        // Signals end of execution
        func done() -> None;
    }
    
    // Create a `robot` (callable instance of interface)
    robot = RobotManipulationInterface();
    """
)

API_DOCS = (
    """
    You can invoke a given function on the robot by calling 
    `robot.<func>("object name"). For example: 
    `robot.set_designated_area("recycling bin")`.

    The API also enables multiple function invocations (separated 
    by newlines).

    Note that each call to `relocate` and `cleanup` *must* be preceded 
    by a call to `set_designated` to be valid!

    To terminate execution for a given action, call `robot.done()`.
    """
)
\end{lstlisting}

In addition to this API definition, we provide three in-context examples in the prompt, as follows:

\begin{lstlisting}[language=Python, caption=In-Context Examples for Real-Robot Code Generation, label=lst:cap-prompt-icl]
ICL_INSTRUCTION = (
    """
    Here are some examples of translating language instructions to 
    API calls. Each instruction defines two variables:
    
    1) a list of interactable `Objects: ["obj1", "obj2", ...]` 
    --> these should be the only "object" arguments to the
    `relocate` and `cleanup` API calls!
    
    2) a mapping of objects to receptacles `Receptacles: 
       {"obj": {"relocate": "<receptacle>", "cleanup": "<receptacle>"}}` 
       --> these should be the only "receptacle" arguments for the 
       `set_designated` API calls!

    Note that if there is *not* a good API call that reflects the 
    desired behavior, it is ok to skip!
    """
)

EXAMPLE_ONE = (
    """
    Instruction: "Retrieve all the crayons and organize them 
    tidily in the designated container."
    Objects: ["crayons", "colored pencils", "notebook", "eraser", 
              "crumpled up napkin"]
    Receptacles: {
      "crayons": {"relocate": "art box", "cleanup": "trash"}, 
      "notebook": {"relocate": "desk", "cleanup": "recycling"}, 
      "eraser": {"relocate": "art box", "cleanup": "trash"}, 
      "crumpled up napkin": {"relocate": "desk", "cleanup": "trash"}
    }
    Program:
    ```
    robot.set_designated("art box");
    robot.relocate("crayons");
    robot.done();
    ```
    """
)

EXAMPLE_TWO = (
    """
    Instruction: "Throw away the half-eaten apple."
    Objects: ["apple", "orange", "half-eaten peach", 
              "coffee cup", "pink plate"]
    Receptacles: {
      "apple": {"relocate": "counter", "cleanup": "trash"}, 
      "orange": {"relocate": "counter", "cleanup": "trash"}, 
      "half-eaten peach": {"relocate": "counter", "cleanup": "trash"}, 
      "coffee cup": {"relocate": "counter", "cleanup": "recycling"}, 
      "pink plate": {"relocate": "counter", "cleanup": "sink"}
    }
    Program:
    ```
    robot.set_designated("trash can");
    robot.cleanup("apple");
    robot.done();
    ```
    """
)

EXAMPLE_THREE = (
    """
    Instruction: "Leave the castle as is in a designated area, then 
                  put away the removeable parts in a continer."
    Objects: ["toy castle", "castle parts", "figurine", "cheerios"]
    Receptacles: {
      "toy castle": {"relocate": "shelf", "cleanup": "toy box"}, 
      "castle parts": {"relocate": "play mat", "cleanup": "toy box"}, 
      "figurine": {"relocate": "shelf", "cleanup": "toy box"}, 
      "cheerios": {"relocate": "play mat", "cleanup": "trash"}
    }
    Program:
    ```
    robot.leave_alone("toy castle");
    robot.set_designated("toy box");
    robot.cleanup("castle parts");
    robot.done();
    ```
    """
)
\end{lstlisting}

Finally, we produce the following continuation string that we use to seed the LLM completion; the $\{, \}$ denote variables that are filled in programmatically at runtime:

\begin{lstlisting}[language=Python, caption=Continuation Prompt for Code Generation, label=lst:cap-prompt-continuation]
CONTINUATION = (
    "
    Instruction: {instruction}
    Objects: {objects}
    Receptacles: {receptacles}
    Program:
    ```
    " + 
    """< LLM CONTINUATION >"""
)
\end{lstlisting}

\subsection{Examples}
\label{appx:examples}

We list several examples our dataset and framework below. 

\begin{figure*}
    \centering
    \includegraphics[width=.95\textwidth]{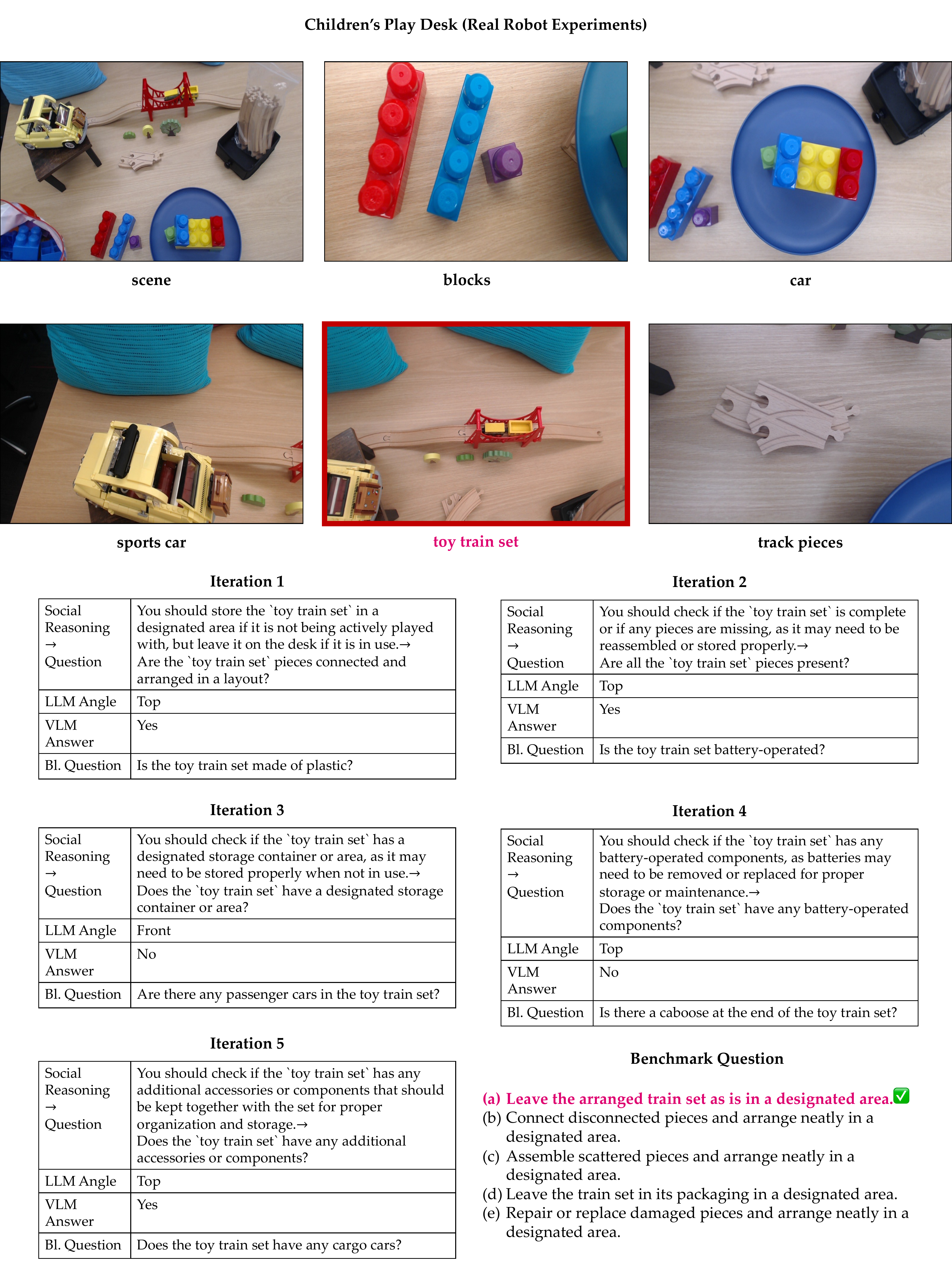}
    \caption{\textbf{Children's Play Desk (Real-Robot Experiments).} Example photos of each object (from a top-down angle). We also provide examples of our framework's outputs for the \emph{toy train set} for all $5$ iterations using InstructBLIP.}
    \label{fig:childrens_play_desk}
\end{figure*}

%% TODO: figure out how to cut better!
\stop

\begin{figure*}
    \centering
    \includegraphics[width=.95\textwidth]{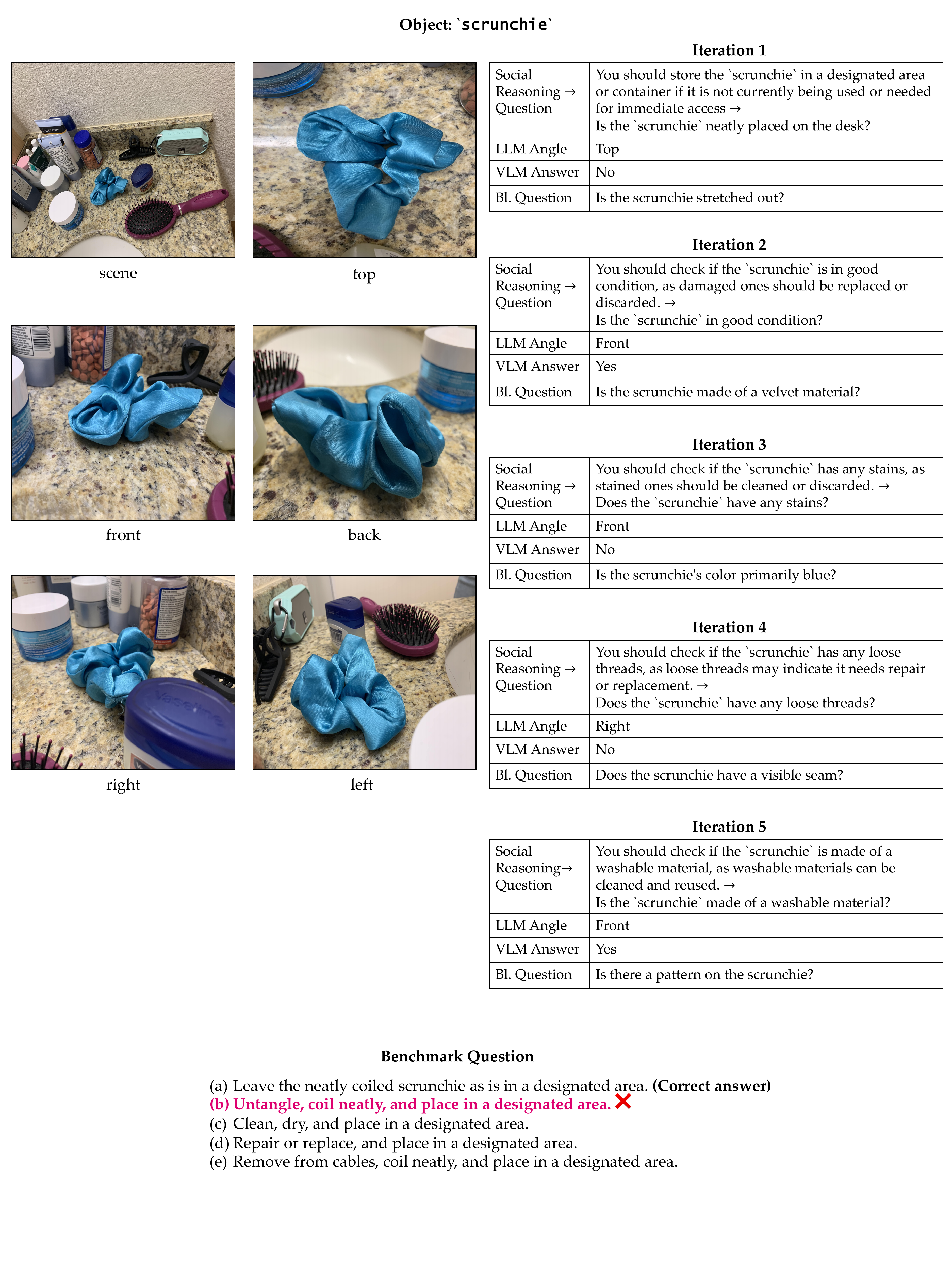}
    \caption{\textbf{Example 1 from \dataset{}.} Scene image and close-up photos of a \emph{scrunchie}. We also provide examples of our framework's outputs for all $5$ iterations using InstructBLIP.}
    \label{fig:example1}
\end{figure*}

\begin{figure*}
    \centering
    \includegraphics[width=.95\textwidth]{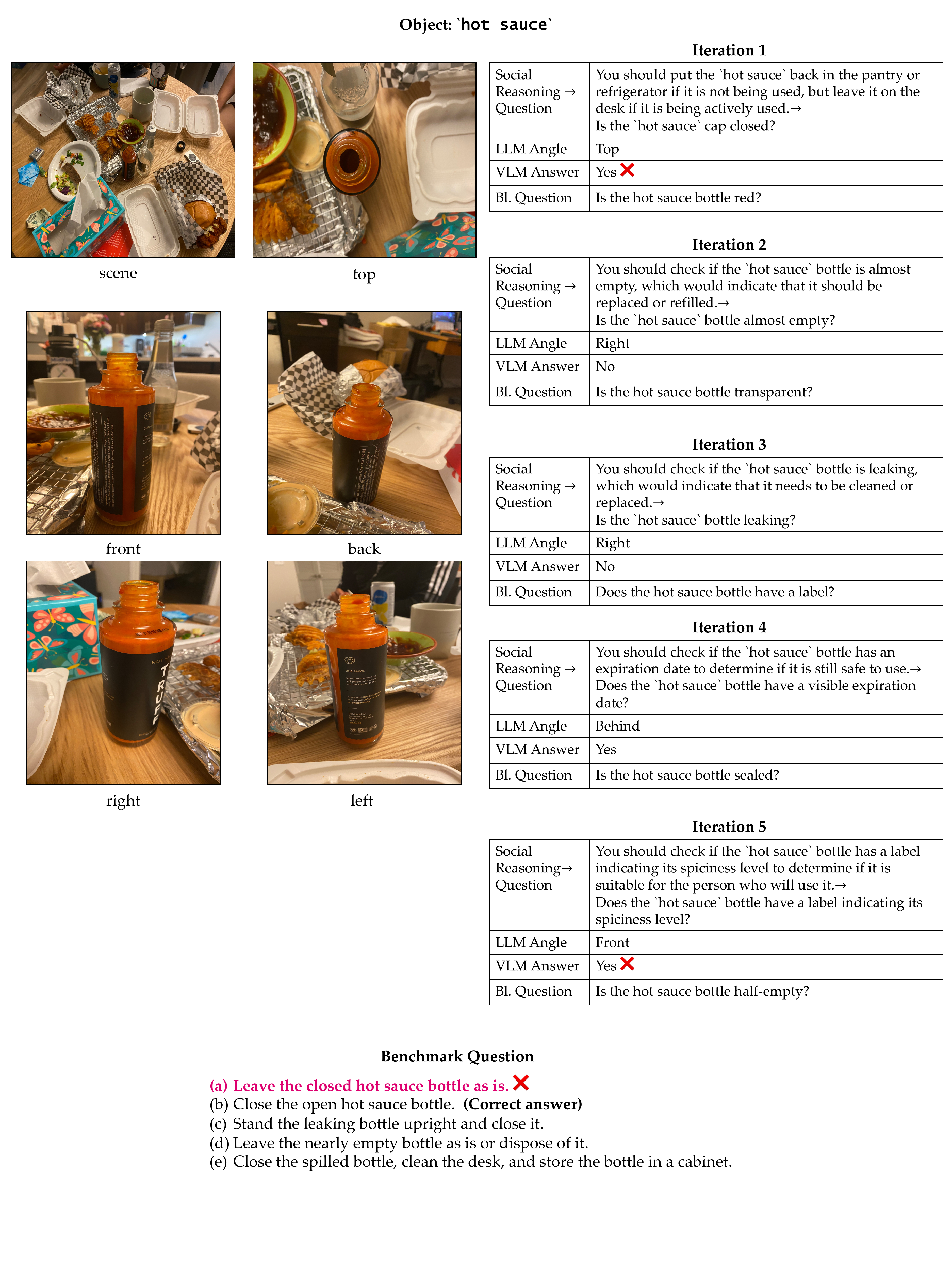}
    \caption{\textbf{Example 2 from \dataset{}.} Scene image and close-up photos of a \emph{hot sauce}. We also provide examples of our framework's outputs for all $5$ iterations using InstructBLIP.}
    \label{fig:example2}
\end{figure*}

\begin{figure*}
    \centering
    \includegraphics[width=.95\textwidth]{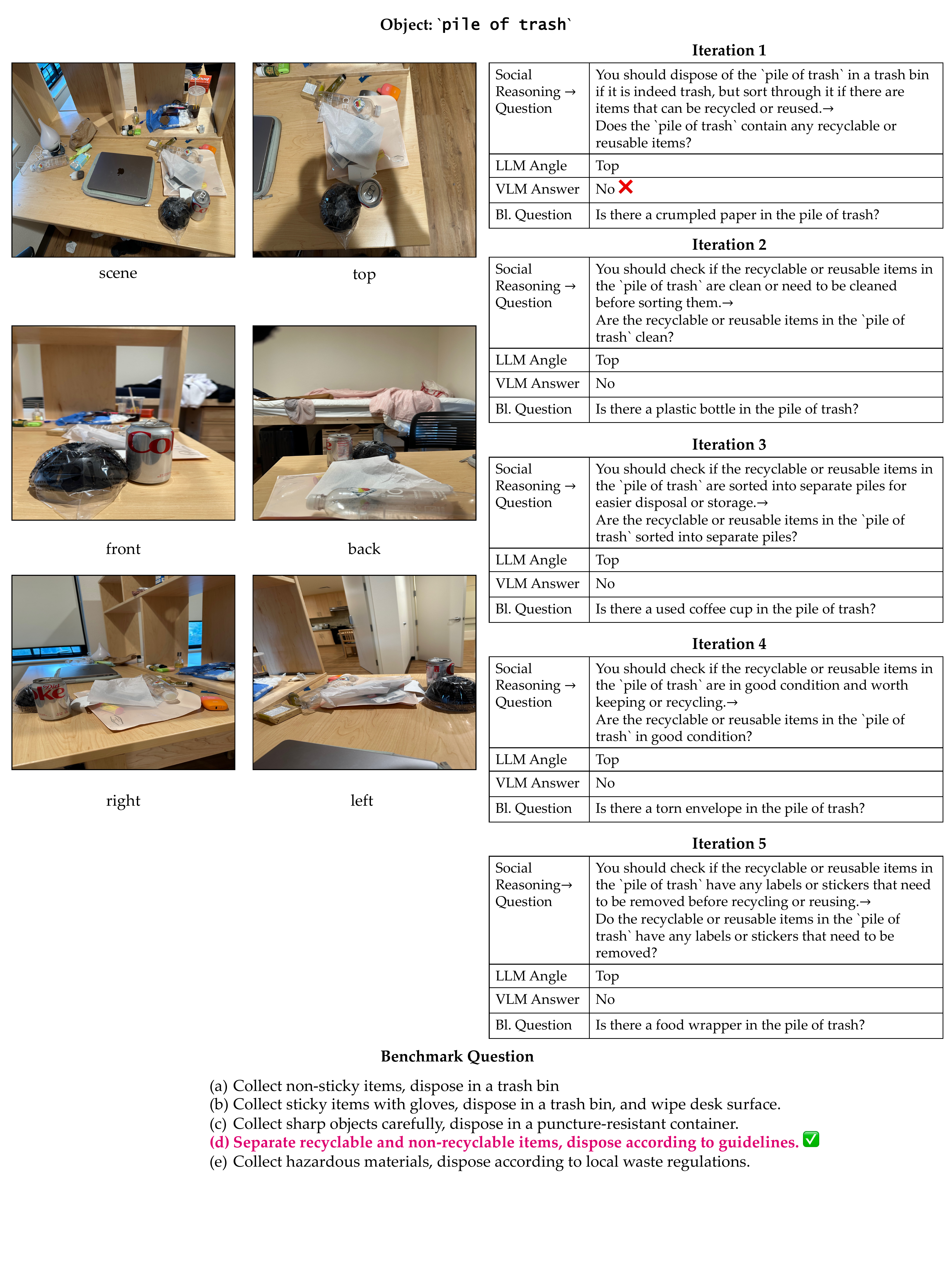}
    \caption{\textbf{Example 3 from \dataset{}.} Scene image and close-up photos of a \emph{pile of trash}. We also provide examples of our framework's outputs for all $5$ iterations using InstructBLIP.}
    \label{fig:example3}
\end{figure*}

\begin{figure*}
    \centering
    \includegraphics[width=.95\textwidth]{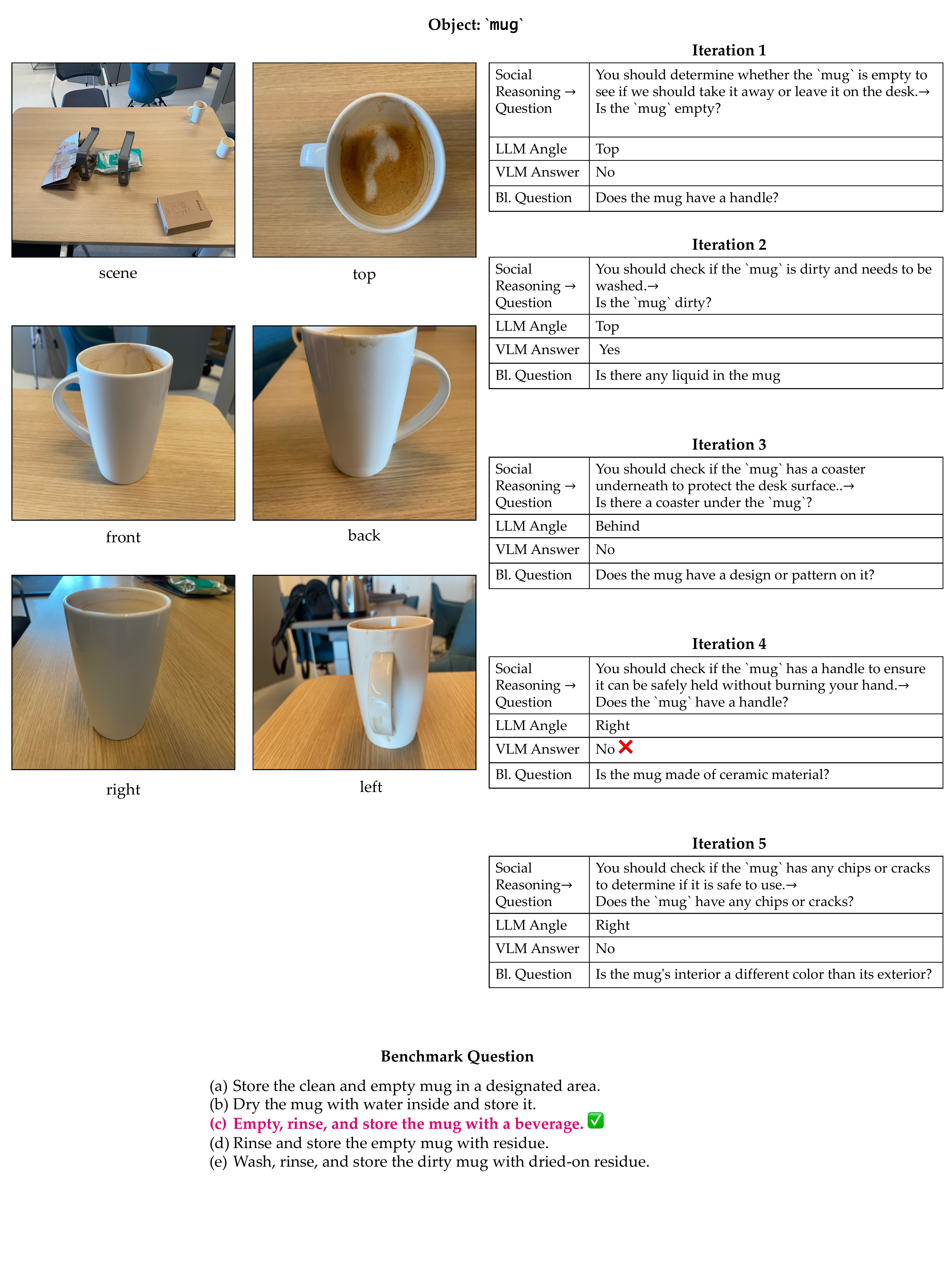}
    \caption{\textbf{Example 4 from \dataset{}.} Scene image and close-up photos of a \emph{mug}. We also provide examples of our framework's outputs for all $5$ iterations using InstructBLIP.}
    \label{fig:example4}
\end{figure*}

\end{document}